\documentclass[10pt,twocolumn,letterpaper]{article}

\usepackage[pagenumbers]{wacv}

\usepackage{siunitx}

\usepackage{graphicx}
\usepackage{xcolor}
\usepackage{soul}
\sethlcolor{yellow}
\usepackage{colortbl}
\usepackage{tikz}
\usepackage{caption}
\usepackage{placeins}
\usepackage[export]{adjustbox}

\renewcommand{\arraystretch}{0.5}
\usepackage{multirow}

\newsavebox{\cropbox}
\newlength{\cropwd}
\newlength{\trimwd}
\newcommand{\cropLRtoBox}[3]{
  \sbox{\cropbox}{\includegraphics[height=#2]{#3}}%
  \setlength{\cropwd}{\wd\cropbox}%
  \ifdim\cropwd>#1
    \setlength{\trimwd}{0.5\dimexpr\cropwd-#1\relax}%
    \raisebox{0pt}[#2][0pt]{%
      \makebox[#1][c]{\clipbox{\trimwd 0pt \trimwd 0pt}{\usebox{\cropbox}}}%
    }%
  \else
    \raisebox{0pt}[#2][0pt]{%
      \makebox[#1][c]{\usebox{\cropbox}}%
    }%
  \fi
}

\usepackage{graphicx}
\usepackage{xcolor}

\newlength{\imgwidth}
\newlength{\imgheight}
\newlength{\imgw}
\newlength{\imgh}
\newlength{\gapSmall}
\newlength{\gapBig}
\newlength{\pairw}
\newlength{\fullw}
\newcommand{\instrpair}[1]{%
  \multicolumn{2}{@{}p{\pairw}@{}}{%
    \centering
    \colorbox{black!5}{%
      \parbox{\dimexpr\linewidth-2\fboxsep\relax}{%
        \centering\scriptsize\sffamily\textbf{Instruction:} #1%
      }%
    }%
  }%
}

\newcommand{\instrwide}[1]{%
  \multicolumn{4}{@{}p{\fullw}@{}}{%
    \centering
    \colorbox{black!5}{%
      \parbox{\dimexpr\linewidth-2\fboxsep\relax}{%
        \centering\scriptsize\sffamily\textbf{Instruction:} #1%
      }%
    }%
  }%
}

\newcommand{\coverH}[3]{
  \sbox{\cropbox}{\includegraphics[height=#2]{#3}}%
  \setlength{\cropwd}{\wd\cropbox}%
  \ifdim\cropwd>#1
    \setlength{\trimwd}{0.5\dimexpr\cropwd-#1\relax}%
    \clipbox{\trimwd 0pt \trimwd 0pt}{\usebox{\cropbox}}%
  \else
    \makebox[#1][c]{\usebox{\cropbox}}%
  \fi
}

\definecolor{wacvblue}{rgb}{0.21,0.49,0.74}
\usepackage[pagebackref,breaklinks,colorlinks,allcolors=wacvblue]{hyperref}

\usepackage{tcolorbox}
\tcbuselibrary{skins,breakable}
\usepackage{etoolbox}

\definecolor{VIBEAccent}{HTML}{333333}      
\definecolor{VIBEAccentLight}{HTML}{f8f9fa}  

\newtcolorbox{vibeabstractbox}{
  enhanced,
  breakable,
  colback=VIBEAccentLight,
  colframe=VIBEAccent,
  frame hidden,
  boxrule=0pt,
  borderline west={3pt}{0pt}{VIBEAccent}, 
  arc=2mm,
  outer arc=2mm,
  left=12pt, right=12pt, top=14pt, bottom=12pt,
  title=Abstract,
  fonttitle=\bfseries\sffamily\small,
  coltitle=white,
  attach boxed title to top left={xshift=10pt, yshift=-11pt}, 
  boxed title style={
    colback=VIBEAccent, 
    frame hidden,
    boxrule=0pt,
    arc=3pt,  
    outer arc=3pt,
    left=5pt, right=5pt, top=2pt, bottom=2pt,
    drop fuzzy shadow=black!30
  }
}
\renewenvironment{abstract}{\begin{vibeabstractbox}}{\end{vibeabstractbox}}

\newtcolorbox{observation}[1][]{
  enhanced,
  breakable,
  colback=VIBEAccentLight, 
  colframe=VIBEAccent,     
  frame hidden,
  boxrule=0pt,
  borderline west={3pt}{0pt}{VIBEAccent}, 
  arc=2mm,
  outer arc=2mm,
  left=12pt, right=12pt, top=14pt, bottom=12pt,
  title={Observation #1},  
  fonttitle=\bfseries\sffamily\small,
  coltitle=white,
  attach boxed title to top left={xshift=10pt, yshift=-11pt}, 
  boxed title style={
    colback=VIBEAccent, 
    frame hidden,
    boxrule=0pt,
    arc=3pt,  
    outer arc=3pt,
    left=5pt, right=5pt, top=2pt, bottom=2pt,
    drop fuzzy shadow=black!30
  }
}

\title{VIBE: Visual Instruction Based Editor}

\author{%
    Grigorii Alekseenko\textsuperscript{*} \hspace{0.1em}
    Aleksandr Gordeev\textsuperscript{*} \hspace{0.1em}
    Irina Tolstykh\textsuperscript{*} \enspace 
    Bulat Suleimanov \hspace{0.1em}
    Vladimir Dokholyan \\ [0.4em]
    Georgii Fedorov \hspace{0.1em}
    Sergey Yakubson \hspace{0.1em}
    Aleksandra Tsybina \hspace{0.1em}
    Mikhail Chernyshov \hspace{0.1em}
    Maksim Kuprashevich\textsuperscript{\textdagger} \\[0.5em]
    \large R\&D Department, SALUTEDEV \\[0.5em]
    {\footnotesize \textsuperscript{*}Equal contribution \qquad \textsuperscript{\textdagger}\texttt{Corresponding author}}
}

\makeatletter
\patchcmd{\@maketitle}{\null}{\null\enlargethispage*{0.85in}\vspace*{-0.85in}}{}{}

\patchcmd{\@maketitle}
  {\iftoggle{wacvrebuttal}{\vspace*{-.3in}}{\vskip .375in}}
  {\iftoggle{wacvrebuttal}{\vspace*{-.3in}}{\vskip 0.15in}}{}{}
\patchcmd{\@maketitle}{\vspace*{24pt}}{\vspace*{14pt}}{}{}
\patchcmd{\@maketitle}{\vskip .5em}{\vskip 0.2em}{}{}
\patchcmd{\@maketitle}{\vspace*{12pt}}{\vspace*{6pt}}{}{}

\apptocmd{\@maketitle}{%
  \vspace{-1.8em}%
  \noindent\begin{abstract}
Instruction-based image editing is among the fastest developing areas in generative AI. Over the past year, the field has reached a new level, with dozens of open-source models released alongside highly capable commercial systems.

However, only a limited number of open-source approaches currently achieve real-world quality. In addition, diffusion backbones, the dominant choice for these pipelines, are often large and computationally expensive for many deployments and research settings, with widely used variants typically containing 6B to 20B parameters.

This paper presents a compact, high-throughput instruction-based image editing pipeline that uses a modern 2B-parameter Qwen3-VL model to guide the editing process and the 1.6B-parameter diffusion model Sana1.5 for image generation. Our design decisions across architecture, data processing, training configuration, and evaluation target low-cost inference and strict source consistency while maintaining high quality across the major edit categories feasible at this scale.

Evaluated on the ImgEdit and GEdit benchmarks, the proposed method matches or exceeds the performance of substantially heavier baselines, including models with several times as many parameters and higher inference cost, and is particularly strong on edits that require preserving the input image, such as an attribute adjustment, object removal, background edits, and targeted replacement. The model fits within 24 GB of GPU memory and generates edited images at up to 2K resolution in approximately 4 seconds on an NVIDIA H100 in BF16, without additional inference optimizations or distillation. Project page: \href{https://riko0.github.io/VIBE/}{https://riko0.github.io/VIBE/}
\end{abstract}\par
  \vspace{-1.25em}%
  \begin{center}
    \includegraphics[
      width=0.90\textwidth,
      height=0.56\textheight,
    ]{figures/collages/vibe\_collage\_main\_final.pdf}
    \captionof{figure}{Illustrative examples of image edits generated by VIBE.}
  \end{center}%
  \vspace{-1.35em}%
}{}{}
\makeatother

\begin{document}

\maketitle

\begin{figure*}[p]
  \centering
  \includegraphics[width=\textwidth,height=0.98\textheight,keepaspectratio]{figures/collages/vibe\_collage\_nonhuman.pdf}
  \label{fig:Collage-nonhuman}
  \caption{Illustrative examples of image edits generated by VIBE.}
\end{figure*}

\begin{figure*}[p]
  \centering
  \includegraphics[width=\textwidth]{figures/collages/vibe\_collage\_human.pdf}
  \label{fig:Collage-human}
  \caption{Illustrative examples of image edits generated by VIBE.}
\end{figure*}
\clearpage

\section{Introduction}
\label{section:introduction}

Instruction-based image editing models allow visual content to be modified according to natural-language commands and promise to democratize content creation. Compared to traditional retouching tools, which require substantial expertise, such generative models offer intuitive, language-based interfaces that are accessible to non-experts. Consequently, instruction-guided editing has become one of the most active directions in generative AI.

Recent proprietary systems have demonstrated rapid progress, including Google Nano Banana Pro \cite{nano_banana_pro} (Gemini 3 Pro Image \cite{google2025nanobananaapi}), OpenAI's GPT Image 1.5 \cite{openai2025chatgptimages} \cite{openai2025gptimage15docs}, and Black Forest Labs' FLUX.1 Kontext models \cite{flux_kontext}. In contrast, open-source research generally trails in both quality and usability. Most open models remain large (6B to 20B parameters) and expensive to train and iterate on, which slows experimentation and limits accessibility \cite{liu2025step1xedit}.

Many practical systems start from a pretrained text-to-image diffusion backbone and adapt it to instruction-based editing. Under this setting, diffusion-based editing is shaped by three design axes: (i) how the reference image is injected, (ii) how the instruction is interpreted, and (iii) how the training pipeline is constructed. 

For reference-image guidance, two common families are (a) channel-wise concatenation of reference latents or features \cite{brooks2023instructpix2pix} and (b) tokenizing visual content and feeding it through the model as part of the input sequence \cite{liu2025step1xedit}. 

For textual guidance, a key architectural choice is whether to rely mainly on the diffusion backbone's native text conditioning \cite{flux_kontext}, or to add an external model that rewrites, expands, or structures the edit intent before conditioning the generator \cite{fu2023mgie}. Many widely used text-to-image diffusion backbones are optimized for text-conditioned generation and therefore rely on text-only conditioning modules (e.g., CLIP \cite{radford2021clip}, T5 \cite{raffel2020t5}, or even an LLM as in Sana1.5 \cite{xie2024sana}). In such pipelines, the conditioning module cannot observe the source image, so it cannot interpret the instruction in the context of the reference content. For image editing, this joint interpretation is often essential. The model must ground the request in what is actually in the input image to resolve ambiguity and preserve source-faithful details. We therefore use an instruction-tuned VLM that ingests both the instruction and the source image and produces a clearer, image-aware conditioning signal for the diffusion.

Since the diffusion backbone still expects conditioning in the representation space of its native text encoder, an additional design decision is the connector that maps the VLM representations into the diffusion model's conditioning space \cite{fu2023mgie,liu2025step1xedit}.

This paper investigates these architectural questions under strict efficiency constraints. We target low-cost inference by combining computationally efficient channel-wise concatenation with a learnable meta-tokens \mbox{mechanism}~\cite{pan2025metaqueries}.

We train with a four-stage pipeline:
\begin{itemize}
    \item \textbf{Alignment}: adapting a VLM to interface with the latent diffusion space via a text-to-image objective on high-aesthetic samples.
    \item \textbf{Pre-training}: learning core editing capabilities by adding image-to-image tasks on large-scale, relatively noisy data.
    \item \textbf{Supervised Fine-Tuning}: carefully tuning on clean and diverse triplets.
    \item \textbf{Direct Preference Optimization (DPO)} \cite{wallace2024diffusion}: aligning the model using high-quality preference data with real-world instructions.
\end{itemize}

The proposed pipeline is flexible and can be applied to other LLM/VLM and diffusion backbones. It also supports backbones that rely on relatively lightweight text encoders, such as the CLIP text encoder \cite{radford2021clip}, because the alignment stage explicitly bridges the language model and the diffusion latent space.

Another focus of our approach is to adopt a model for real-world challenges, rather than for technical benchmarks. We focus on real user requests and curate or synthesize instructions that better match human phrasing than templated or purely LLM-generated prompts.

The data collected for this pipeline spans diverse sources and is optimized for low noise and in-the-wild distributions. We combine specialist-model pipelines, distilled signals from both open and proprietary editing systems, autonomous triplet-mining pipelines, filtered open-source image editing and computer vision datasets, manually collected tripod-captured photographs, and additional sources. We also apply extensive augmentation, in particular, the pipeline relies heavily on triplet inversion and bootstrapping, which reduces data cost in both compute and annotation.

Historically, different instruction-guided image editing methods assume different tolerances for unintended modifications to the source image, including the degree to which pixel-level appearance, scene composition, subject identity, and other attributes must be preserved. In this work, we target strict source consistency: any change not explicitly requested by the instruction is treated as an issue and addressed throughout all stages of training and evaluation. This objective is particularly challenging for edit categories that intrinsically encourage global transformations, such as style transfer.

To maintain dataset quality, we use a multi-stage filtering framework, including learned triplet scoring via a fine-tuned Gemini-based validator and auxiliary checks such as face-embedding constraints for identity preservation and image-quality scoring to prevent quality degradation.

In summary, our primary contributions are:
\begin{enumerate}
    \item We present an open-source, ultra-fast, compact instruction-guided image editing system trained on $\approx15$ million triplets, based on Qwen3-VL-2B-Instruct \cite{qwen3vl2025} and the Sana1.5-1.6B diffusion model \cite{xie2024sana}.
    \item We propose a flexible four-stage training pipeline that can be adapted to different diffusion backbones and LLM/VLM front-ends, enabled by our architectural choices.
    \item We provide results, analysis, and insights covering experimental design, data collection, augmentation, and filtering, along with ablation studies.
\end{enumerate}

\section{Related Works}
\label{section:related_work}
Instruction-based image editing has rapidly evolved, with progress driven by innovations in model architectures, guidance mechanisms, and training strategies. Early methods were often training-free, operating directly on pretrained diffusion models via inversion or attention control
\cite{hertz2022prompt, mokady2023null, cao2023masactrl, tumanyan2023plug, feng2025dit4edit}.
While cost-efficient, these approaches struggle to achieve high-quality results.
As a result, the field has shifted toward training-based paradigms that fine-tune diffusion backbones on large-scale triplets
\cite{brooks2023instructpix2pix, zhang2023magicbrush, hui2024hq, ge2024seed, zhao2024ultraedit, yu2025anyedit}.
Interestingly, many widely used training triplets were bootstrapped with earlier editing systems, underscoring the tight coupling between scalable data generation and model progress
\cite{brooks2023instructpix2pix, zhang2023magicbrush, zhao2024ultraedit}.

\subsection{Production-oriented open editors and efficiency constraints.}
Despite rapid progress, production-level editing quality remains concentrated in a limited number of systems.
Recent open foundation editors increasingly unify text-to-image generation and instruction-based editing within a single model family, but often rely on relatively large diffusion backbones: ranging from 6B to 20B parameters in recent releases (e.g., LongCat-Image/Z-Image at 6B, FLUX.1 Kontext \texttt{[dev]} at 12B, and Qwen-Image-Edit built on a 20B Qwen-Image backbone) \cite{longcat2025image, zimage2025, hf_flux_kontext_dev, qwen_image_edit}.
Such scale raises both training and inference cost: it slows development iteration (ablations, retraining/fine-tuning, and production updates) and increases user-facing latency and cost per edit, reducing the number of interactive refinement cycles a user can afford before reaching the desired result.
Motivated by these costs, recent work has begun to study more compute-efficient diffusion transformers and training recipes, including Sana-style backbones \cite{xie2024sana}.
In this work, we focus on the same efficiency-first setting and pair a compact 2B-class VLM with a 1.6B diffusion backbone to deliver low-latency, low-cost edits with strict source consistency.

\subsection{Architectures for Conditioning the Source \mbox{Image}}
A core design choice in diffusion-based editing is how to condition the denoising process on the source image.
A widely used and computationally efficient approach is \emph{channel-wise concatenation}, introduced by InstructPix2Pix \cite{brooks2023instructpix2pix},
where the source-image latent is concatenated with the noisy latent along the channel dimension.
This design keeps inference lightweight and is often favored in latency-sensitive settings.

Another family uses \emph{token-wise} multimodal conditioning, where visual content is tokenized and injected through attention as part of the model input sequence.
This enables richer interactions between the source image, the instruction, and intermediate representations throughout the network \cite{liu2025step1xedit, flux_kontext},
but often comes with higher architectural and computational overhead.
Recent foundation editors further popularize single-stream diffusion transformers that process text and image tokens in a unified sequence,
and report strong editing behavior as part of a general generation-and-editing capability \cite{qwen_image, longcat2025image, zimage2025}.
In contrast, we retain the practical efficiency of channel-wise conditioning while relying on compact VLM guidance and data/recipe choices to reach production-level behavior under tight deployment constraints.

\subsection{Architectures for Interpreting Instructions}
Another major axis is how the textual instruction is represented and grounded in the source image.
Many editors rely primarily on the diffusion backbone's native text conditioning and improve instruction following through data scaling and training recipes
\cite{brooks2023instructpix2pix, zhang2023magicbrush, mao2025ace++, yu2025anyedit, flux_kontext}.
A complementary line of work introduces a stronger VLM to interpret the instruction in the context of the source image and to produce a clearer edit intent for the generator \cite{fu2023mgie}.
Recent open foundation editors increasingly integrate strong VLM components directly into the editing stack.
For example, Qwen-Image-Edit extends an open image foundation model with multimodal conditioning for instruction-driven edits \cite{qwen_image_edit},
while LongCat-Image-Edit and Z-Image-Edit report dedicated editing variants trained within similarly unified generation-and-editing frameworks \cite{longcat2025image, zimage2025}.
Our pipeline follows the same high-level direction using a modern VLM to guide image editing, but is explicitly optimized for throughput and strict consistency at compact scale.

\subsection{Training Pipelines, Data, and Alignment}
Beyond model architecture, the training pipeline itself is a crucial factor. While early works focused on dataset curation \cite{brooks2023instructpix2pix, zhang2023magicbrush, hui2024hq, ge2024seed, zhao2024ultraedit}, recent research has investigated more sophisticated schemes, including multi-stage training and auxiliary objectives \cite{shi2024seededit, fu2025univg, mao2025ace++, xia2025dreamomni}. A common practical issue in editing fine-tuning is \emph{catastrophic forgetting}, where adapting a pretrained text-to-image model to specialized editing triplets can degrade its original generative prior, harming robustness and aesthetic quality. Another recurring difficulty is \emph{interface alignment}: when a VLM is used to interpret edits, its representations must be mapped into the conditioning space expected by the diffusion backbone, and naive end-to-end training can be unstable or sample-inefficient.

Many recent open-source pipelines refine editing behavior with post-training alignment signals, for example via preference-based objectives (and, in some cases, distillation from stronger teacher editors), to improve perceptual quality and instruction adherence \cite{wallace2024diffusion}. Separately, recent foundation editors emphasize large-scale joint pretraining (often including image-to-image objectives) followed by supervised post-training and alignment as a practical route to strong editing performance \cite{qwen_image, longcat2025image, zimage2025}.

In our four-stage setup, we first perform an \emph{alignment} stage that establishes a VLM-to-diffusion connection by adapting the new VLM and connector to the frozen DiT model's embedding space. This stage uses a text-to-image objective on high-aesthetic data, stabilizing the interface before the model learns editing-specific behaviors. We then introduce large-scale image-to-image pre-training, followed by supervised fine-tuning on curated triplets, and finally apply preference-based post-training (DPO) to improve edit quality and reliability \cite{wallace2024diffusion, rafailov2023dpo}. To maintain real-world behavior, we emphasize aggressive quality control throughout data construction and training, including augmentation (e.g., triplet inversion and bootstrapping) and multi-stage filtering/validation to reduce unintended modifications and enforce strict source consistency.

\subsection{Consistency and Real-World Instruction Distributions}
Instruction-guided editing methods differ substantially in their tolerance for unintended changes to the source image,
including identity preservation, background stability, lighting consistency, and fine-grained appearance control.
Maintaining strict source consistency is especially challenging for edit categories that encourage global transformations (e.g., stylization)
or that require delicate, localized modifications without collateral drift.
Another practical gap is the instruction distribution.
In many academic datasets, instructions are annotator-written or LLM-generated and can differ from real user queries in phrasing, ambiguity, and intent.
While recent datasets and human-feedback efforts improve coverage and quality \cite{HIVE, hui2024hq, zhao2024ultraedit},
matching in-the-wild instruction style remains challenging.
Our work explicitly targets real user behavior by grounding instruction text in real-world queries and filtering aggressively for consistency,
enabling a compact model to behave reliably under realistic prompting.
\begin{figure*}[ht]
    \centering
    \includegraphics[width=\textwidth]{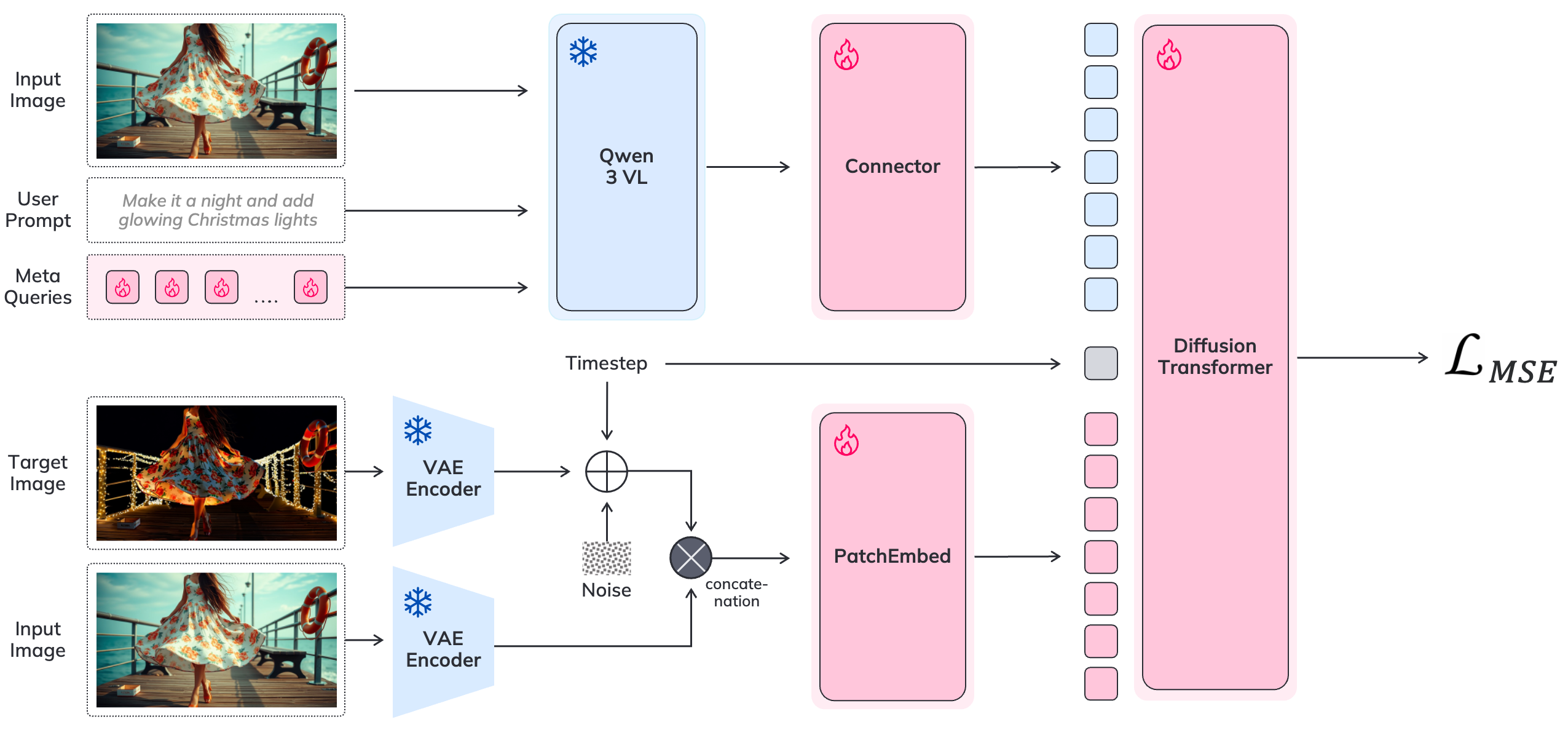}
    \caption{Model architecture.}
    \label{fig:Architecture}
\end{figure*}

\section{Method} 
\label{section:method}

Our architecture integrates two primary components: (i) a Large Vision-Language Model which employs learnable meta tokens (detailed in Section \ref{mllm_based_section}) to interpret the user's instruction within the context of the input image; and (ii) a diffusion transformer that employs a generative process to synthesize the edited image. To bridge these components, we use a connector module designed to align the editing intent with the diffusion model, as detailed in Section~\ref{connector}.
The overall pipeline is illustrated in Figure~\ref{fig:Architecture}.

In this work, we introduce a model that generates images at 2K-class resolutions with diverse aspect ratios. This substantially improves quality in terms of preserving fine-grained details from the source image. 

\subsection{Reference Image Guidance}
\label{ref_image}

To guide the diffusion process with reference image $\mathbf{I}_\text{R}$, we first encode it into latent representation $\mathbf{L}_\text{R}\in\mathbb{R}^{ c\times h\times w}$ utilizing frozen VAE block. 

To integrate $\mathbf{L}_\text{R}$ into the denoising pipeline, we employ \textbf{\mbox{channel-wise} concatenation}. In contrast to \mbox{sequence-wise} concatenation, which concatenates the reference latents along the \emph{token} dimension -- thereby increasing the sequence length and directly increasing the computational cost of attention mechanism -- the channel-wise formulation concatenates the reference latents $\mathbf{L}_\text{R}$ with the noise latents along the \emph{channel} dimension. A widened input convolution then restores the original channel dimensionality and projects the result into token space. This preserves the number of tokens and therefore leaves the attention complexity unchanged, maintaining high generation throughput. 

\subsection{Textual Guidance Based on VLM}
\label{mllm_based_section}
\paragraph{Interface between VLM and Diffusion model}
In \cite{pan2025metaqueries} the authors demonstrate that directly using hidden states from the final layer of an VLM is a suboptimal way to guide a diffusion model. To effectively bridge the modality gap, we randomly initialize special meta-tokens and add them into the VLM's vocabulary while keeping the model weights frozen. The number of tokens is treated as a hyperparameter.   

During the forward pass, meta tokens are concatenated manually with instruction tokens and propagated through all layers of the network. It is noteworthy that we do not rewrite the user instruction into expressive instructions as seen in MGIE~\citep{fu2023mgie}, but do add prompt prefix such as "What would the image look like if \{user instruction\}?". 

\begin{equation}
\mathbf{\hat{T}}_{\text{M}} = \operatorname{VLM}\bigl(\mathbf{I}_\text{R}, \mathbf{U}_\text{I}, \mathbf{T}_{\text{M}}\bigr).
\end{equation}
\noindent Here, $\mathbf{I}_\text{R}$ denotes the reference image, $\mathbf{U}_\text{I}$ is the sequence of instruction tokens, and $\mathbf{\hat{T}}_{\text{M}} \in \mathbb{R}^{N \times d}$ are $N$ learnable meta-token embeddings. The VLM jointly processes the concatenated sequence through all transformer layers and outputs contextualized meta-token hidden states $\mathbf{\hat{T}}_{\text{M}}$.

\subsection{Connector Design}
\label{connector}
The contextualized meta-token hidden states $\mathbf{\hat{T}}_{\text{M}}$ are then mapped to the conditioning space of the diffusion backbone via a lightweight trainable connector module. Concretely, the connector is implemented as a stack of Transformer encoder blocks operating only on the meta-token sequence:
\begin{equation}
\mathbf{C}_\text{T} = \operatorname{Connector}\!\left(\mathbf{\hat{T}}_{\text{M}}\right),
\end{equation}
\noindent where $\mathbf{C}_\text{T}$ denotes the resulting conditioning features used by the diffusion model.

\subsection{Training Approach}
\paragraph{Connector Alignment} 
In our configuration, both the VLM and the diffusion model are initialized from the pretrained checkpoints. Only connector module and meta tokens are randomly initialized and trained from scratch. Consequently, during the initial training phases, the signal transmitted from the VLM to the diffusion model via the connector is significantly distorted. While the weights of the connector are coming to reasonable values, the weights of the pretrained and unfrozen diffusion model undergo significant alterations. This leads to an irreversible degradation of its generative capabilities, thereby reducing the quality of the final output. 

To address this issue, we propose an intermediate preadaptation step for the connector. We freeze the VLM and the diffusion model and train the pipeline exclusively on a text-to-image generation task. Once satisfactory performance metrics are achieved, we consider the connector to be aligned. Subsequently, we proceed to train the model on the primary image editing task.

\begin{observation}[1]
    Incorporating this additional alignment stage not only enhances the quality of the generated images but also improves the model's ability to follow instructions.
\end{observation}

\paragraph{T2I Data Injections} A common practice for training image editing models is to use specialized datasets consisting exclusively of $\langle source\_image, instruction, target\_image\rangle$ triplets. However, we find that this strategy can substantially degrade the model’s foundational text-to-image generation capability. In practice, the model overfits to artifacts in the relatively limited editing data and consequently generalizes poorly to real-world images and user instructions.

To address this issue, we propose a mixed-data training strategy that frames editing as constrained image generation, rather than as plain image-to-image translation. We train the model on instruction-based editing triplets together with a set of high-quality text-to-image pairs. Technically, we mix both data types within each batch. For T2I samples, we feed an empty (black pixels) conditioning image which is masked out in the attention layers. Additionally, we employ task-specific text templates: for T2I, the input is structured as ``\textit{generate the image by description: \{prompt\}}'', while for editing, we use ``\textit{what will this image be like if \{prompt\}}''. This joint training provides two benefits: (i) it regularizes learning and reduces overfitting to the limited, often artificial editing data, and (ii) it preserves the model’s original generative prior by keeping standard text-to-image generation active throughout training.

\begin{observation}[2]
Multi-task training prevents drift from the robust pre-trained initialization, which is crucial for high-fidelity edits that require synthesizing new content (e.g., object addition). Text-to-image data acts as a distributional anchor, keeping the final model both a strong editor and a capable generator for flexible, creative image manipulation.
\end{observation}

\paragraph{Multi-stage training}

To enhance training efficiency following the connector alignment phase (performed at a resolution of $512^2$), we adopt a multi-stage training strategy for the DiT model. The detailed configuration of our training pipeline, including data ratios and resolution strategies, is summarized in Table~\ref{tab:training_stages}.

During the pre-training stage, the model is trained at an average resolution of $1024^2$ with variable aspect ratios. Subsequently, we execute the SFT phase, performed at resolutions up to $2048^2$. In this phase, we utilize a large-scale, high-quality, and strictly filtered dataset (described in Section~\ref{sec:datasets:sft} and Section~\ref{sec:dataset_log_grounded}) comprising both synthetic and real images.

Throughout the pre-training and SFT phases, we jointly optimize the model for two tasks: image editing and T2I generation. During these stages, the learnable meta-tokens, the connector module, and the diffusion model are updated, while the VLM backbone remains frozen. Following these supervised stages, we employ DPO for the DiT model (see Section~\ref{sec:dpo}).

Notably, regarding resolution management, we diverge from traditional progressive resizing \citep{esser2024scaling}. Since we fine-tune a pre-trained diffusion backbone rather than training from scratch, the standard low-resolution warm-up becomes redundant. Instead, we employ a mixed-resolution strategy during both pre-training and SFT, training simultaneously on data spanning resolutions from $384^2$ to $2048^2$ with diverse aspect ratios. This approach yields several key benefits:
\begin{itemize}
    \item It ensures the model preserves its robust high-resolution generative priors while adapting to the editing task.
    \item The simultaneous processing of varied resolutions preserves the diversity of triplets, and allow us to avoid image upscaling, which can harm generation quality.
\end{itemize} 
To implement this efficiently, we utilize adaptive batch sizing. We dynamically adjust the batch size based on input dimensions by increasing the batch size for lower-resolution inputs to ensure full utilization of GPU resources.

\begin{observation}[3]
    Simultaneous multi-resolution training with diverse aspect ratios significantly accelerates convergence and results in superior generation quality compared to iterative resolution increase.
\end{observation}

\subsection{Preference Alignment}
\label{sec:dpo}

\paragraph{Preliminaries}

Diffusion-DPO~\citep{wallace2024diffusion} adapts the Direct Preference Optimization framework~\citep{rafailov2023dpo} to align diffusion models with human preferences. Unlike RLHF\cite{ouyang2022training}, which requires training a separate reward model, DPO optimizes the policy directly using ranked pairs of images $(x_w, x_l)$ conditioned on context $c$.

While standard DPO relies on exact log-likelihoods $\log p_\theta(x|c)$, these are intractable for diffusion models. To address this, Diffusion-DPO approximates the likelihood ratio using the Evidence Lower Bound (ELBO), re-formulating the objective via denoising errors. The loss function is defined as:

\begin{equation}
\label{eq:diff_dpo}
    \mathcal{L}_{\text{Diff-DPO}}(\theta) = - \mathbb{E}_{(x_w, x_l, c), t, \epsilon} \left[ \log \sigma \left( \beta \left( \delta_{\theta}(x_w) - \delta_{\theta}(x_l) \right) \right) \right],
\end{equation}
where $\delta_{\theta}(x)$ represents the implicit reward derived from the difference in reconstruction errors between the reference model ($\epsilon_{\text{ref}}$) and the trained model ($\epsilon_{\theta}$):
\begin{equation}
    \delta_{\theta}(x) = \lVert \epsilon - \epsilon_{\text{ref}}(x_t, t, c) \rVert_2^2 - \lVert \epsilon - \epsilon_{\theta}(x_t, t, c) \rVert_2^2
\end{equation}

Here, $x_t$ denotes the noisy latent at timestep $t$, $\epsilon$ is the added noise, and $\beta$ is a regularization hyperparameter. Intuitively, this objective encourages the model to minimize the denoising error for the preferred image $x_w$ relative to the reference model, while effectively increasing it for the disfavored image $x_l$.

\paragraph{Post-training}
During the post-training phase, we employ Diffusion-DPO to align the model with human preferences. Specifically, we utilize DPO to address two primary challenges: (i) visual artifacts that arise during real image editing, and (ii) failures in instruction adherence. The detailed process for dataset construction is provided in Section~\ref{section:dpo_dataset}.

In the context of multi-reward optimization, targeting both instruction adherence and aesthetic quality, we eschew scalarization techniques, such as weighted sums or geometric means, to define the preference direction. These rigid formulations often fail to reconcile the inherent inconsistencies between conflicting objectives; for instance, optimizing aggressively for image aesthetics may inadvertently degrade the model's faithfulness to the editing instructions. Consequently, relying on a single aggregated score as a proxy for utility risks reward over-optimization and results in unbalanced alignment.

While recent approaches, such as DreamBoothDPO~\cite{ayupov2025dreamboothdpo} and CaPO~\cite{lee2025calibrated}, introduce complex sampling strategies to navigate the multi-preference distribution, we explore a more direct avenue to achieve Pareto optimality. We adopt a \textit{strict dominance} strategy for preference pair construction: during training, we select a pair $(x_w, x_l)$ only if the preferred sample $x_w$ strictly outperforms the rejected sample $x_l$ across \textit{both} reward criteria simultaneously.
\begin{observation}[4]
Strict-dominance pair filtering reduced reward over-optimization and produced more balanced gains than scalarized objectives. In our experiments, it matched or outperformed more involved multi-preference sampling strategies.
\end{observation}

\subsection{Implementation Details}

We employ the Qwen3-VL-2B\footnote{\url{https://huggingface.co/Qwen/Qwen3-VL-2B-Instruct}}~\citep{qwen3vl2025} model as the VLM backbone, which produces hidden states with an embedding dimension of 2048. For our text-to-image generation backbone, we use the Sana1.5-1.6B model\footnote{\url{https://huggingface.co/Efficient-Large-Model/SANA1.5_1.6B_1024px}}\citep{xie2025sana}. We utilize 224 learnable meta tokens and the connector consists of 4 Transformer encoder blocks, with these hyperparameters selected through extensive empirical experimentation.

\begin{table*}[t]
\centering
\caption{
\textbf{Training stages of the proposed architecture.} 
The pipeline consists of an initial alignment of the connector, followed by multi-stage training of the diffusion backbone (DiT).
The columns \textbf{Edit (\%)} and \textbf{T2I (\%)} denote the data sampling ratio between editing triplets and text-to-image pairs.
Note that the VLM backbone remains frozen throughout the entire process.
}
\label{tab:training_stages}
\resizebox{\textwidth}{!}{%
\begin{tabular}{l c c c c l}
\toprule
\multirow{2}{*}{\textbf{Training Stage}} & \multirow{2}{*}{\textbf{Resolution}} & \multirow{2}{*}{\textbf{Trainable Modules}} & \multicolumn{2}{c}{\textbf{Data Ratio}} & \multirow{2}{*}{\textbf{Data Composition}} \\
\cmidrule(lr){4-5}
& & & \textbf{T2I (\%)} & \textbf{Edit (\%)} & \\
\midrule
\textbf{I. Connector Alignment} & $512^2$ & Connector, Meta Tokens & 100\% & 0\% & Text-to-Image pairs \\
\cmidrule(lr){1-6}
\textbf{II. Pre-training} & $\le 1024^2$ & \multirow{2}{*}{\shortstack{DiT, Connector, \\ Meta Tokens}} & 68\% & 32\% & Editing triplets + T2I data \\[2pt]
\textbf{III. SFT} & $\le 2048^2$ & & 34\% & 62\% & Large-scale high-quality filtered triplets + T2I \\[2pt]
\cmidrule(lr){1-6}
\textbf{IV. Preference Alignment} & $\le 2048^2$ & DiT & 0\% & 100\% & Preference pairs $(x_w, x_l)$ \\
\bottomrule
\end{tabular}%
}
\end{table*}

\section{Assessor}
\label{sec:assessor}

Accurate evaluation of image editing quality remains an open problem, as standard metrics often fail to correlate sufficiently with human perception. In our work, a robust automated metric is essential, serving as the primary tool for filtering training data.

Following the approach in~\cite{nhr}, we developed a specialized assessor. Initially, we fine-tuned a \textbf{Gemini 2.0 Flash} model on a set of \num{4350} examples. Subsequently, we expanded the dataset to \num{12335} examples and trained a non-proprietary \textbf{Qwen-2.5-VL-7B} model utilizing  LoRA ~\cite{lora}. Validation was performed on a held-out set of \num{2994} samples.

Table~\ref{tab:assessor_metrics} presents the performance of our models compared to vanilla baselines. As shown, the fine-tuned models demonstrate significantly higher correlation with human judgments compared to their vanilla counterparts. This confirms that task-specific fine-tuning is essential for establishing a reliable filtering tool.

\begin{table}[ht]
\centering
\caption{Quality metrics of the assessor models on validation data ($N=\num{2994}$). I — Instruction, A — Aesthetic. Here \textbf{(V)} denotes vanilla base models, and \textbf{(F)} indicates fine-tuned models.}
\renewcommand{\arraystretch}{1.05}
\label{tab:assessor_metrics}
\begin{tabular}{lcccc}
\toprule
Model & I MAE $\downarrow$ & I $\rho \uparrow$ & A MAE $\downarrow$ & A $\rho \uparrow$ \\
\midrule
Qwen-2.5-VL-7B (V)      & 1.030 & 0.437 & 0.936 & 0.198 \\
Gemini 2.5 Flash (V)    & 1.040 & 0.452 & 0.862 & 0.289 \\
Gemini 3 Flash (V) & 0.863 & 0.619 & 0.709 & 0.486 \\
\midrule
\textbf{Gemini 2.0 Flash (F)} & 0.687 & 0.649 & 0.601 & 0.496 \\
\textbf{Qwen-2.5-VL-7B (F)}   & \textbf{0.672} & \textbf{0.641} & \textbf{0.551} & \textbf{0.573} \\
\bottomrule
\end{tabular}
\end{table}
\section{Datasets} 
\label{section:datasets}

\paragraph{For pretraining,} mixtures of publicly available large-scale editing datasets, together with synthetic data from perception and recognition datasets, were initially explored, totaling up to 21 million triplets. The goal was to initialize the model for image editing with broad coverage by training on many edit types, scenes, and instruction styles. However, this early-experiment mixed corpus was too noisy and led to degradation in downstream quality. 
\begin{observation}[5]
Despite large-scale, high-quality SFT, we observed persistent negative transfer from noisy pretraining, with artifacts and failure modes introduced early not fully overridden during SFT.
\end{observation}
Using early prototype models and recent open source models, the most diverse dataset was therefore remastered and a smaller but higher-quality subset was selected, totaling $\approx 7.7$ million triplets. This size was still large enough to maintain diversity, but it was close to the size of the SFT dataset, so SFT could shape the final behavior, while pretraining still added broad instruction and content diversity.

\paragraph{For T2I,} an additional 48 million aesthetically curated images from multiple T2I datasets were assembled. These images or subsets were used during both pretraining and SFT to improve the model's ability to generate visually appealing content.

\paragraph{For SFT,} $\approx 6.8$ million high-quality triplets from diverse sources were used, including inverted samples and compositional bootstrapping.

\paragraph{For DPO,} a specially designed Generation-Augmented Retrieval-based dataset with \num{176532} highest-quality triplets and real-world instructions was used.

Summary can be seen in Table~\ref{table:datasets}.

\subsection{Pretraining}
\label{section:datasets:pretrain:self_data}

\begin{figure*}[t]
\centering
\setlength{\tabcolsep}{0pt}
\renewcommand{\arraystretch}{1}

\newlength{\gap}
\setlength{\gap}{8pt}
\newlength{\colw}
\setlength{\colw}{\dimexpr(\textwidth-3\gap)/4\relax}
\setlength{\pairw}{\dimexpr2\colw+\gap\relax}

\makebox[\textwidth][l]{%
\begin{tabular}{@{}c@{\hspace{\gap}}c@{\hspace{\gap}}c@{\hspace{\gap}}c@{}}

\includegraphics[width=\colw,height=0.205\textheight]{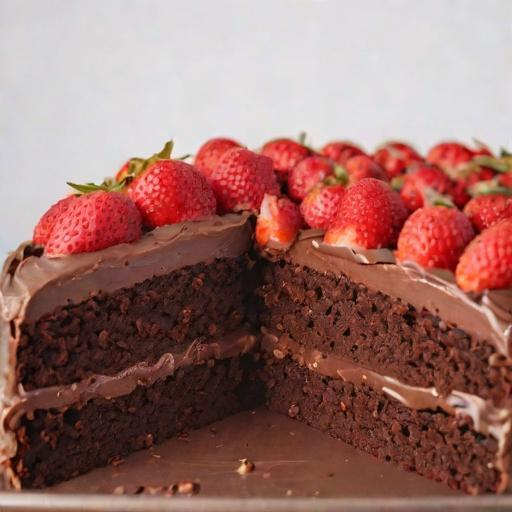} &
\includegraphics[width=\colw,height=0.205\textheight]{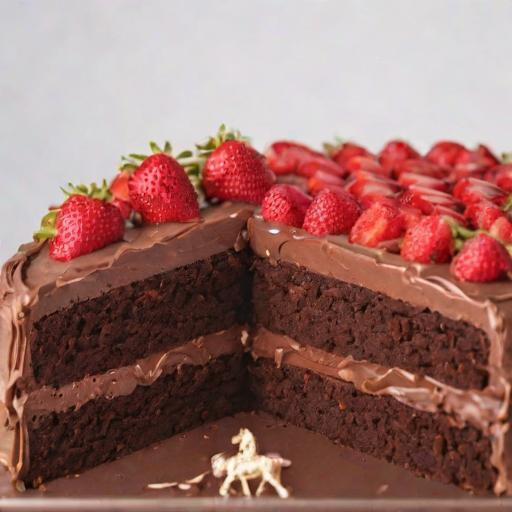} &
\includegraphics[width=\colw,height=0.205\textheight]{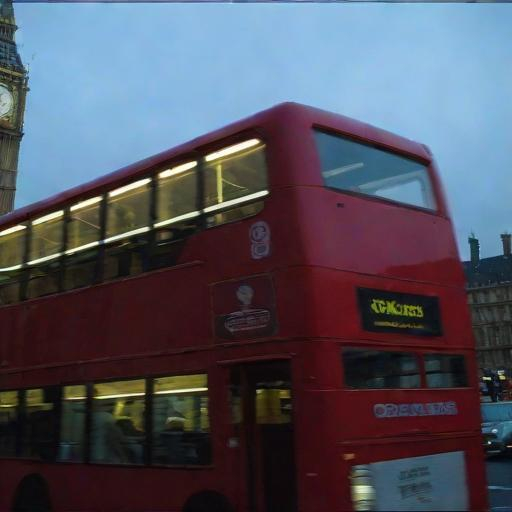} &
\includegraphics[width=\colw,height=0.205\textheight]{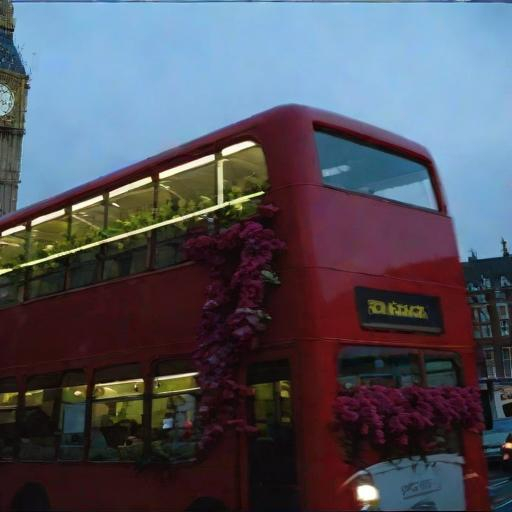}
\\[8pt]

\includegraphics[width=\colw,height=0.285\textheight]{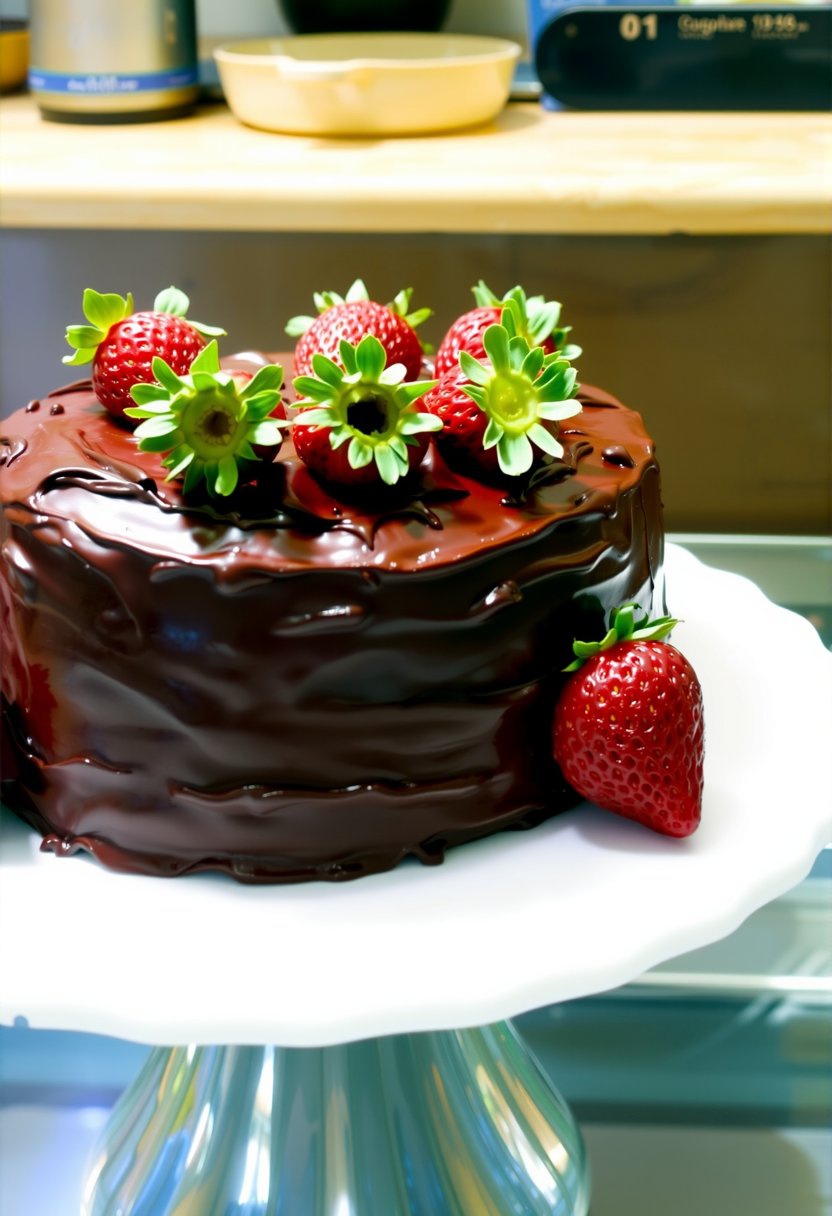} &
\includegraphics[width=\colw,height=0.285\textheight]{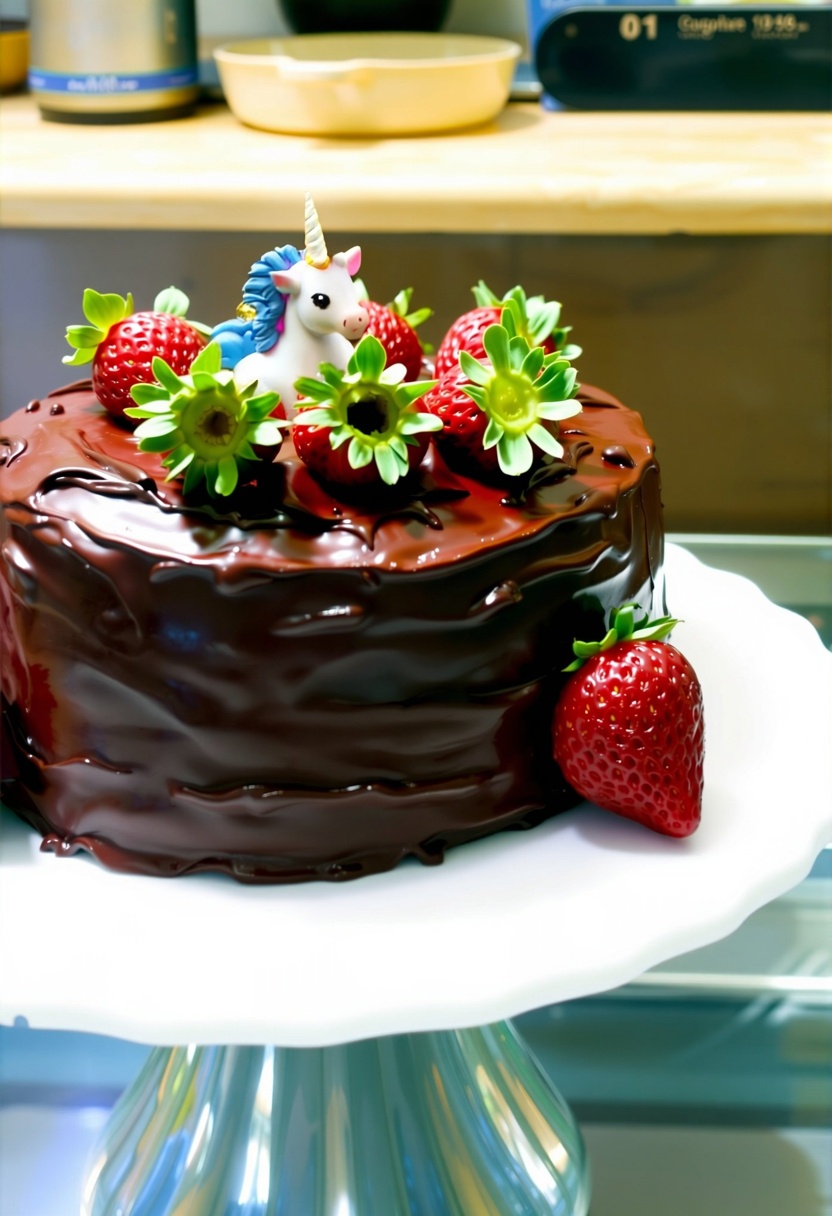} &
\includegraphics[width=\colw,height=0.285\textheight]{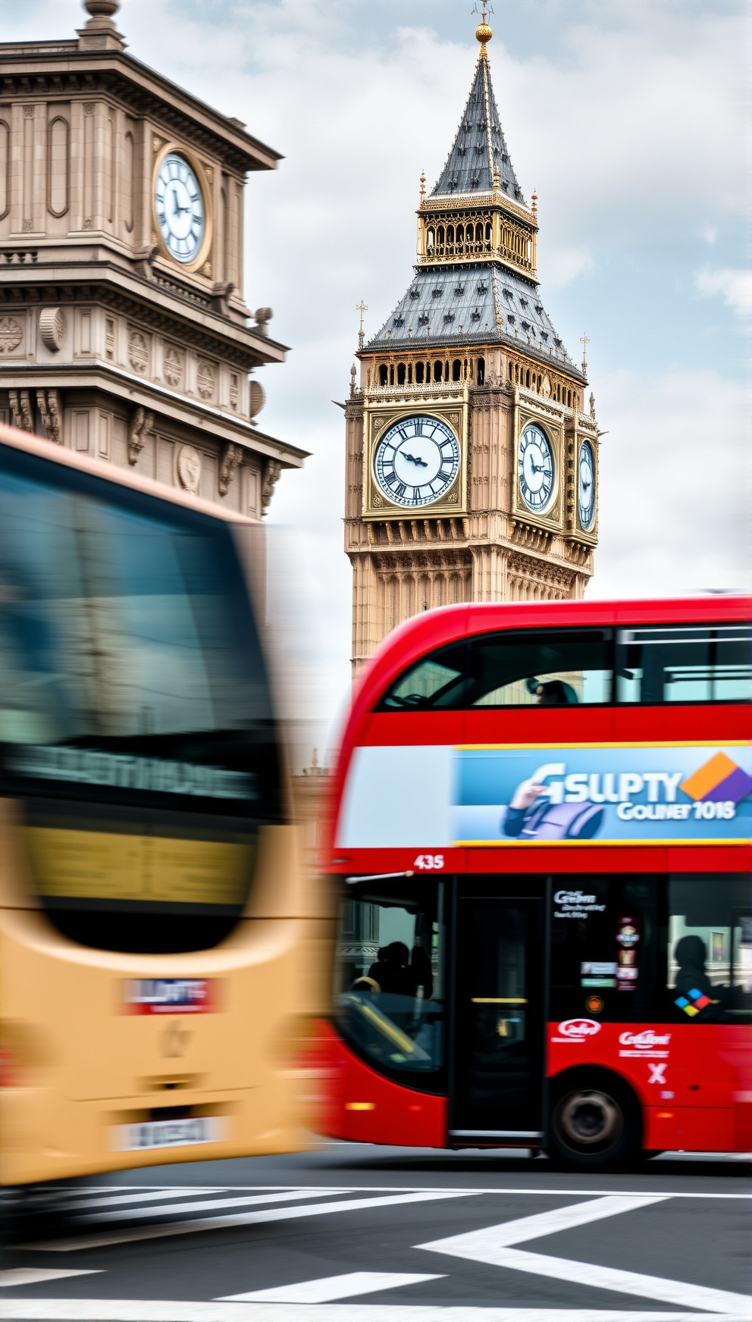} &
\includegraphics[width=\colw,height=0.285\textheight]{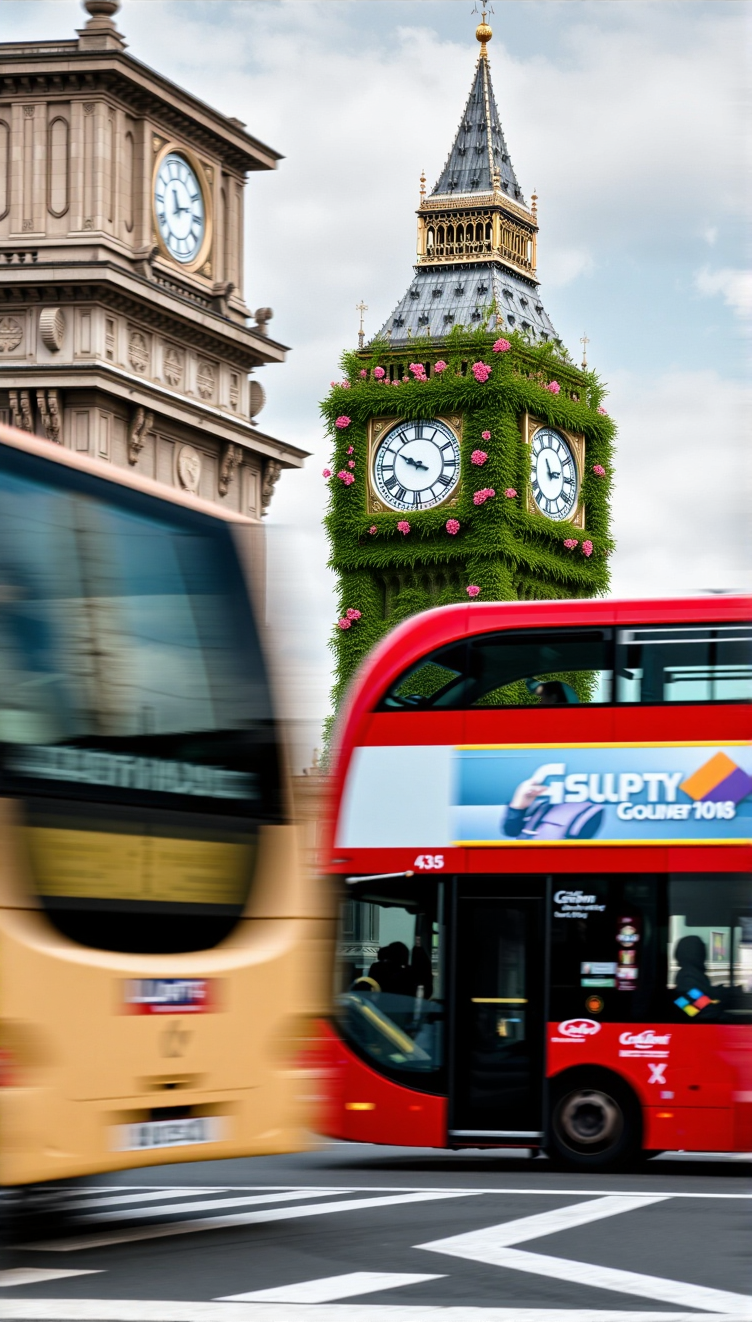}
\\[4pt]

\multicolumn{2}{@{}p{\pairw}@{\hspace{\gap}}}{%
  \centering
  \colorbox{black!5}{%
    \parbox[t]{\dimexpr\linewidth-2\fboxsep\relax}{%
      \centering\scriptsize\sffamily\textbf{Instruction:} Add a unicorn to the scene%
    }%
  }%
} &
\multicolumn{2}{@{}p{\pairw}@{}}{%
  \centering
  \colorbox{black!5}{%
    \parbox[t]{\dimexpr\linewidth-2\fboxsep\relax}{%
      \centering\scriptsize\sffamily\textbf{Instruction:}
      Turn Big Ben into a clock tower covered in vines and flowers%
    }%
  }%
}

\end{tabular}%
}

\vspace{-2pt}
\caption{\small
Examples of UltraEdit. Top row is the original set, bottom row is the remastered version.
}
\label{fig:ultra_edit_remake_dense_13}
\end{figure*}

\paragraph{UltraEdit Remake}
In early experiments, the strongest pretraining results were observed when using UltraEdit as a basis for extension among other large-scale datasets, due to its diversity. At the same time, the original UltraEdit images were low resolution (512$\times$512) and all images were square, which was a problem for our multi-resolution training. Overall noise was also extremely high due to different types of issues (see Figure~\ref{fig:ultra_edit_remake_dense_13}, first row). Eventually, despite very good diversity, this dataset had the lowest overall quality among all large-scale datasets, as shown in~\cite{nhr}. Because it includes text captions of source images, the images were regenerated with proprietary and internal models. Higher resolutions were sampled randomly from the range $[860, 2200]$ for each dimension, with the aspect ratio restricted to $[1{:}6, 6{:}1]$. Prompt adherence and content consistency were validated with Qwen2-VL~\cite{qwen_vl}.

Then, the automated self-mining pipeline initially described in~\cite{nhr} was applied, excluding the Gemini-based validation stage~\cite{gemini} to reduce cost at the pretraining scale. Conceptually, this pipeline over-generates multiple candidate edits for each $\langle source\_image, instruction\rangle$ pair using an instruction-guided image editing model, and then filters or ranks candidates with a validator to retain only high-fidelity $\langle source\_image, instruction, edited\_image \rangle$ triplets.

Given the dataset size, a retry strategy was used for this dataset: candidates were generated until one passed all checks or 5 attempts were exhausted. In total, \num{6420724} triplets were obtained, including the same inversion described in~\cite{nhr}. See Figure~\ref{fig:ultra_edit_remake_dense_13} for examples.

\begin{figure*}[t]
\centering
\setlength{\tabcolsep}{0pt}
\renewcommand{\arraystretch}{1}

\setlength{\gap}{8pt}
\setlength{\colw}{\dimexpr(\textwidth-\gap)/2\relax}
\setlength{\fullw}{\dimexpr2\colw+\gap\relax}

\makebox[\textwidth][c]{%
\begin{tabular}{@{}c@{\hspace{\gap}}c@{}}

\includegraphics[width=\colw,height=0.285\textheight]{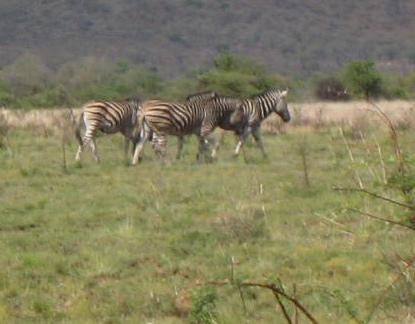} &
\includegraphics[width=\colw,height=0.285\textheight]{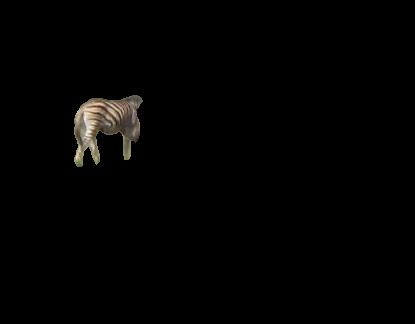}
\\[4pt]

\multicolumn{2}{@{}p{\fullw}@{}}{%
  \centering
  \colorbox{black!5}{%
    \parbox[t]{\dimexpr\linewidth-2\fboxsep\relax}{%
      \centering\scriptsize\sffamily\textbf{Instruction:} Preserve exclusively the left zebra.%
    }%
  }%
}

\end{tabular}%
}

\caption{Example of background removal on the LVIS dataset. High-quality dataset annotations and carefully crafted engineering heuristics enable automatic instruction generation, making localization and object pointing somewhat tricky.}
\label{fig:lvis_back_example}
\end{figure*}

\begin{figure*}[t]
\centering
\setlength{\tabcolsep}{0pt}

\makebox[\textwidth][c]{%
\scalebox{1.02}[1.04]{%
\begin{tabular}{@{}c@{\hspace{\gapSmall}}c@{\hspace{\gapBig}}c@{\hspace{\gapSmall}}c@{}}
  \includegraphics[width=\imgw,height=\imgh]{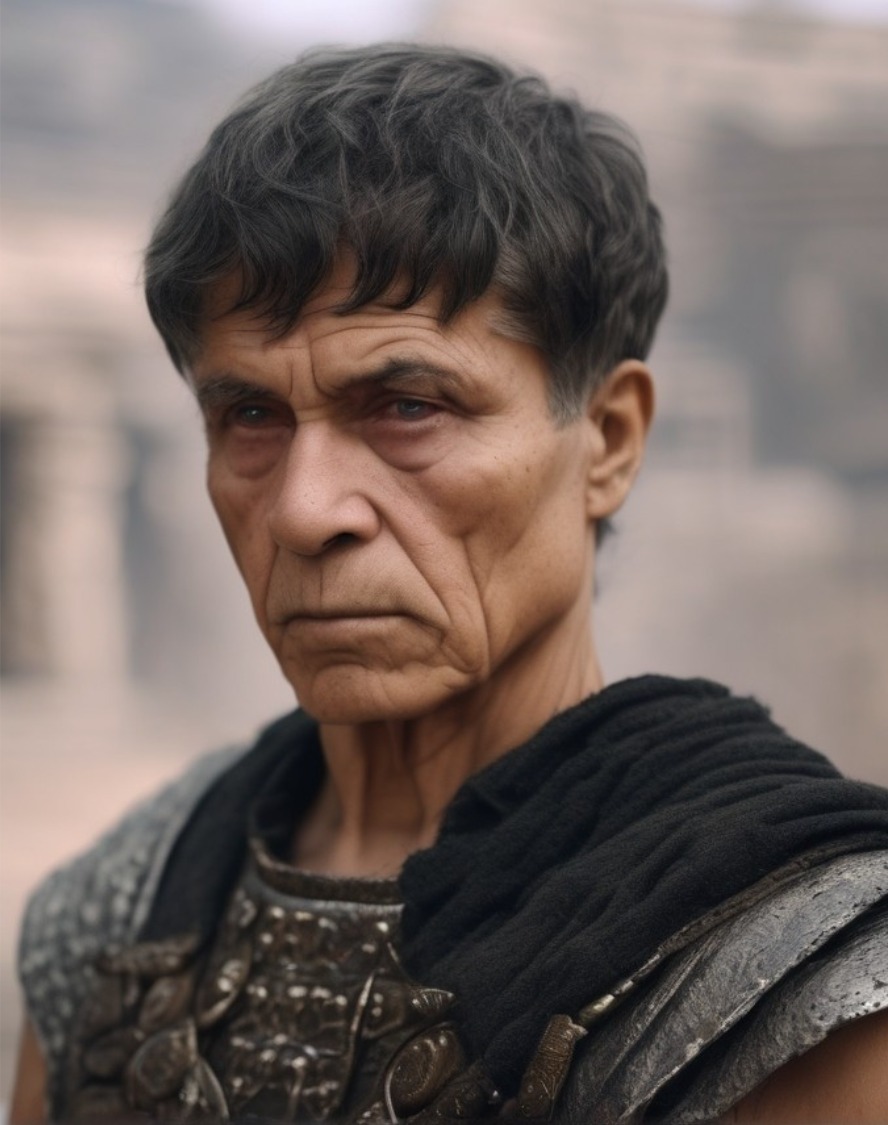} &
  \includegraphics[width=\imgw,height=\imgh]{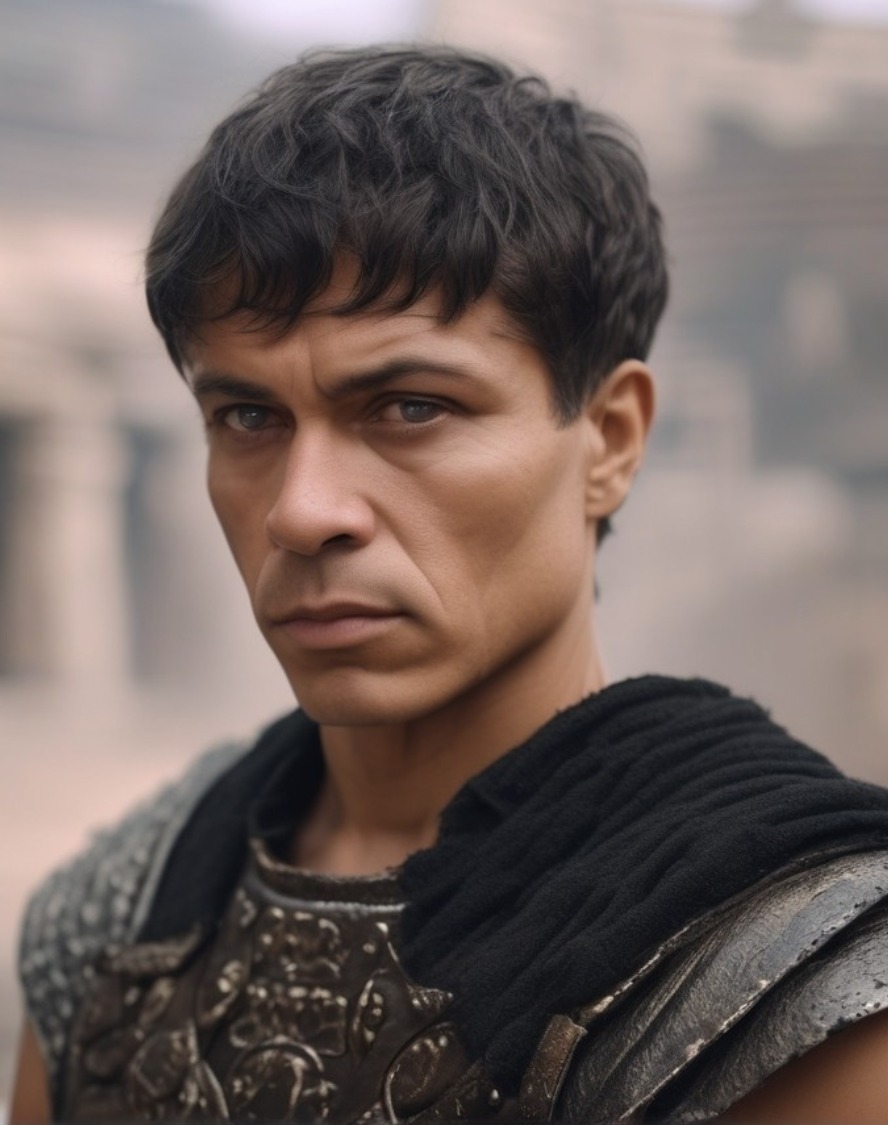} &
  \includegraphics[
    width=\imgw,
    height=\imgh,
    viewport=128 0 1152 1280,
    clip
  ]{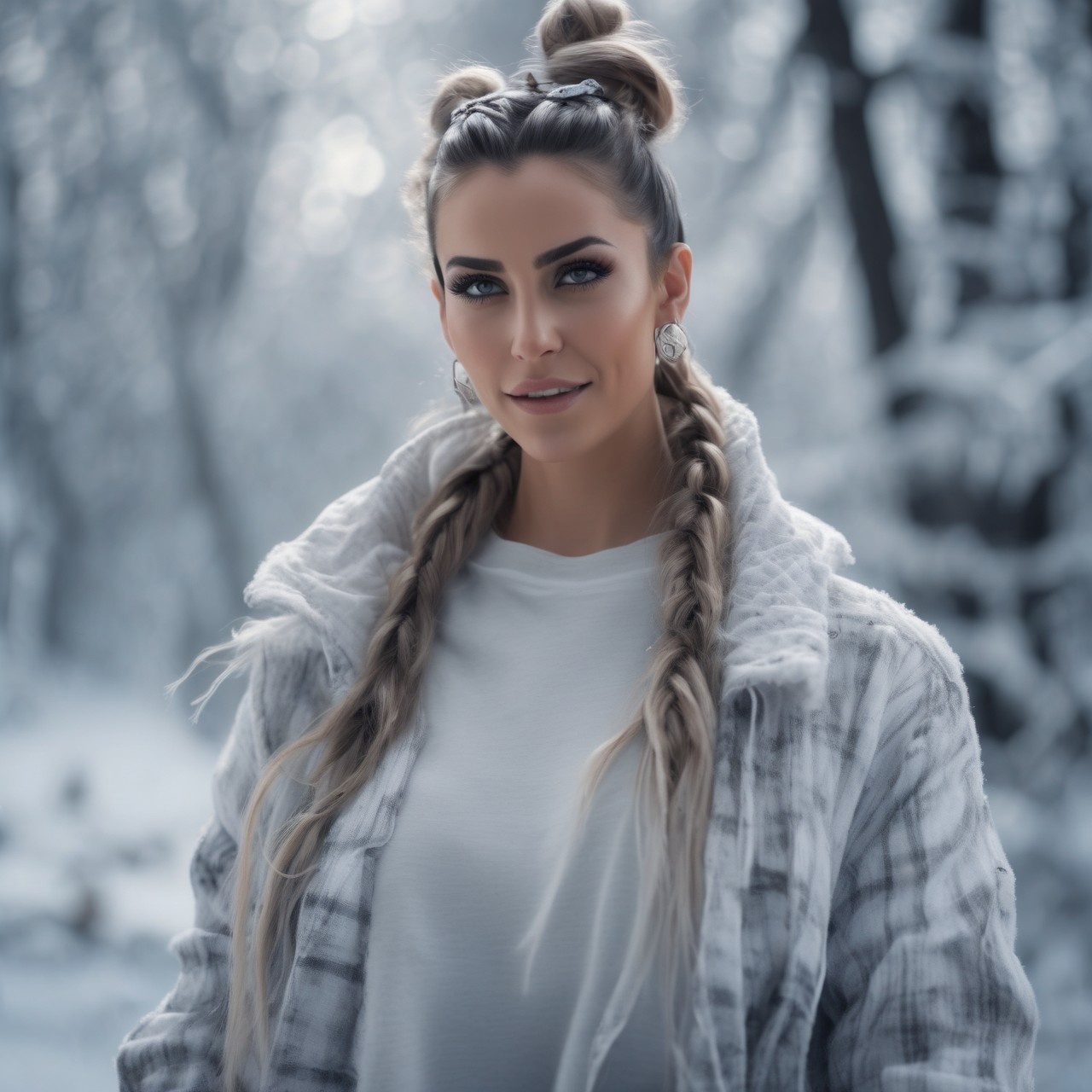} &
  \includegraphics[
    width=\imgw,
    height=\imgh,
    viewport=128 0 1152 1280,
    clip
  ]{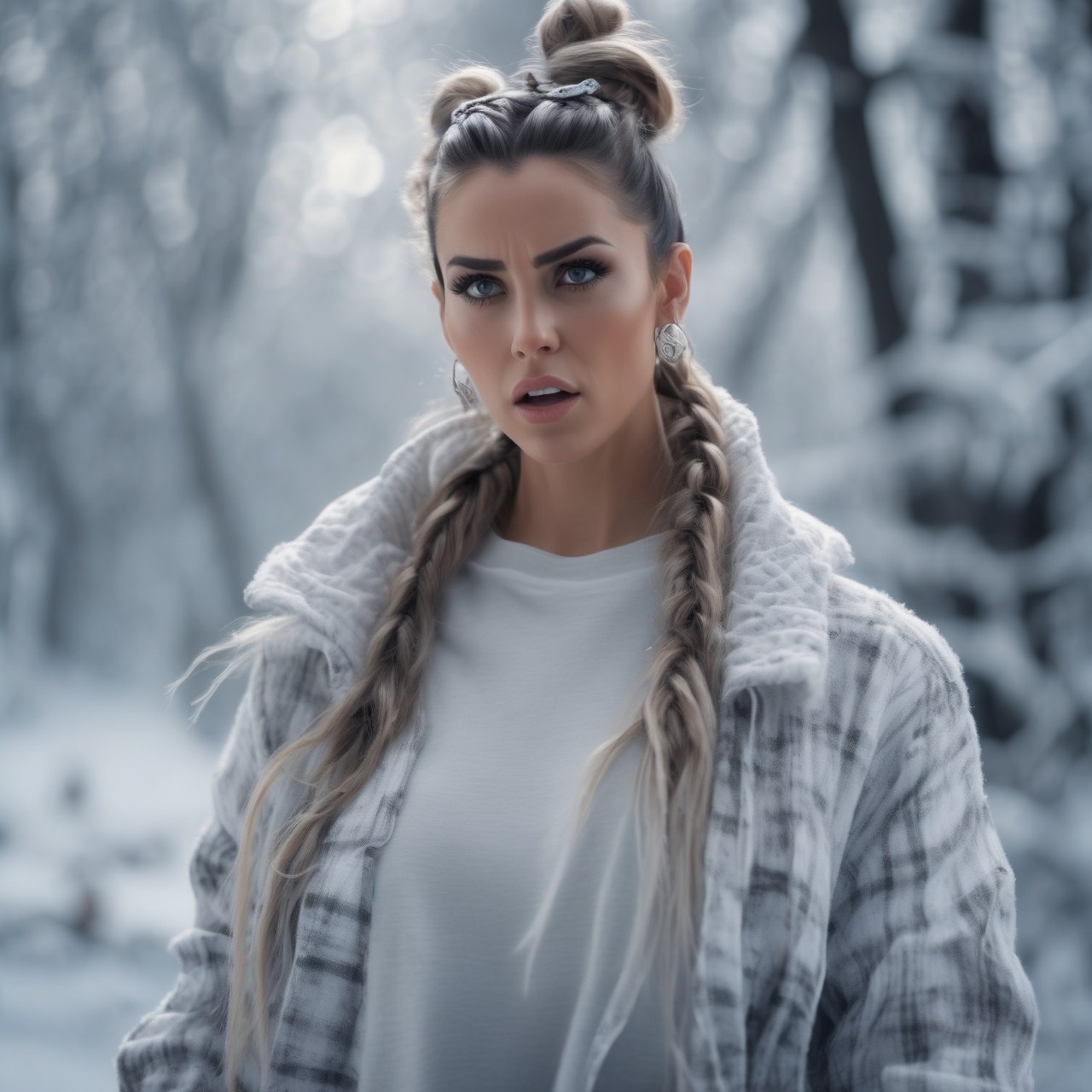} \\
  
  \noalign{\vspace{2pt}}
  \instrpair{Make him 30 years younger} &
  \instrpair{Go from smile to an intense surprise} \\
  \noalign{\vspace{8pt}}

  \multicolumn{2}{@{}p{\pairw}@{}}{%
    \centering
    \includegraphics[
      width=\linewidth,
      height=\imgh,
      viewport=48 0 976 576,
      clip
    ]{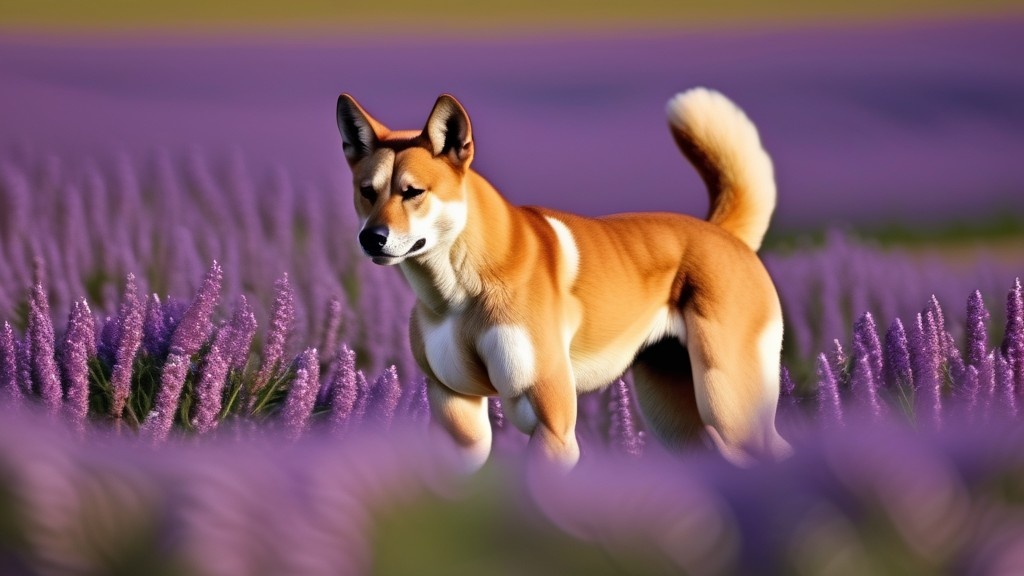}%
  } &
  \multicolumn{2}{@{}p{\pairw}@{}}{%
    \centering
    \includegraphics[
      width=\linewidth,
      height=\imgh,
      viewport=48 0 976 576,
      clip
    ]{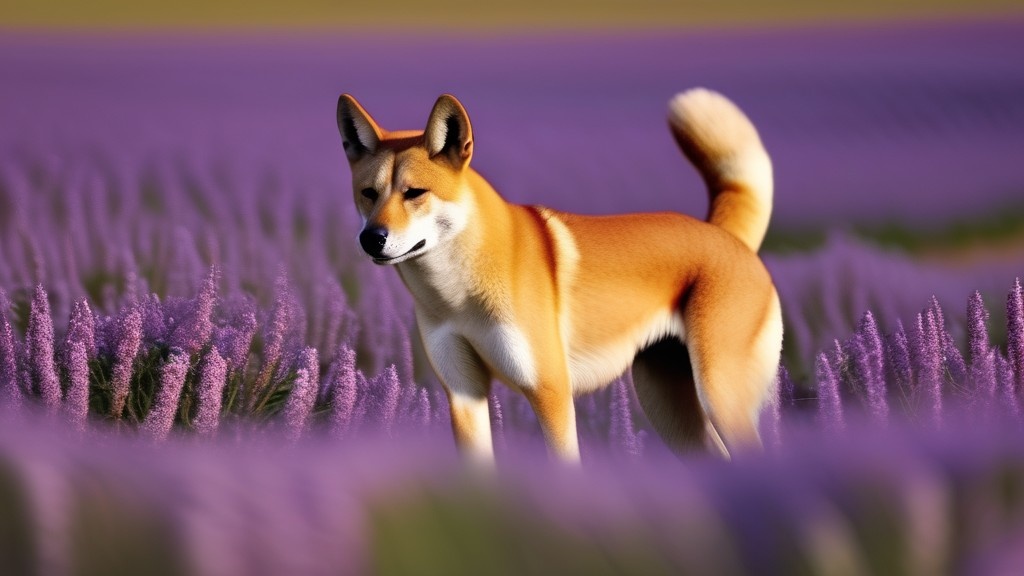}%
  } \\

  \noalign{\vspace{2pt}}
  \instrwide{Make the dingo slim, without muscles}
\end{tabular}%
}%
}

\caption{Examples of Visual Concept Sliders triplets.}
\label{fig:sliders_example}
\end{figure*}

\begin{figure*}[t]
\centering
\setlength{\tabcolsep}{0pt}
\renewcommand{\arraystretch}{1}

\setlength{\colw}{\dimexpr0.5\textwidth-4pt\relax}

\begin{tabular}{@{}c@{\hspace{8pt}}c@{}}
  \includegraphics[
    width=\colw,
    height=0.285\textheight
  ]{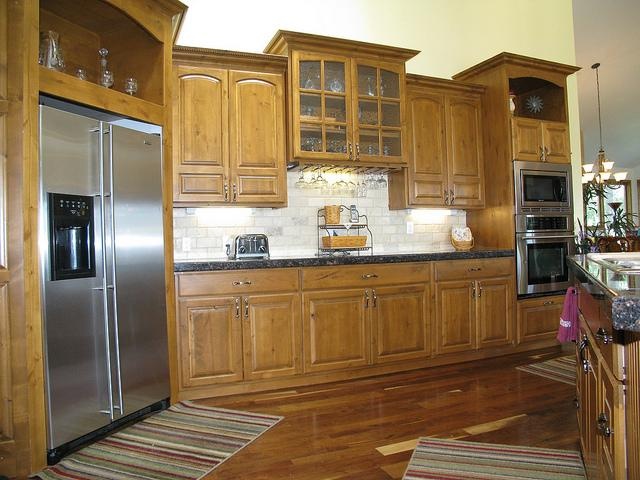} &
  \includegraphics[
    width=\colw,
    height=0.285\textheight
  ]{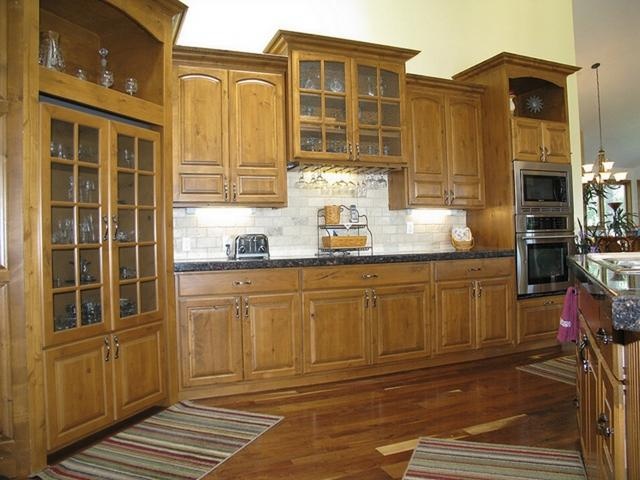} \\
  [4pt]

  \multicolumn{2}{@{}p{\dimexpr2\colw+8pt\relax}@{}}{%
    \centering
    \colorbox{black!5}{%
      \parbox[t]{\dimexpr\linewidth-2\fboxsep\relax}{%
        \centering\scriptsize\sffamily
        \textbf{Instruction (inverted):}
        Swap the modern refrigerator on the left for a large antique wooden hutch with glass doors.%
      }%
    }%
  }%
\end{tabular}

\caption{Example of a triplet obtained with an inpainting model and the LVIS dataset.}
\label{fig:lvis_inpaint}
\end{figure*}

\begin{figure*}[t]
  \centering
  \setlength{\tabcolsep}{0pt}
  \renewcommand{\arraystretch}{1}

  \setlength{\gap}{8pt}
  \setlength{\colw}{\dimexpr(\textwidth-\gap)/2\relax}
  \setlength{\fullw}{\dimexpr2\colw+\gap\relax}

  \begin{tabular}{@{}c@{\hspace{\gap}}c@{}}
    \includegraphics[
      width=\colw,
      height=0.285\textheight,
      keepaspectratio
    ]{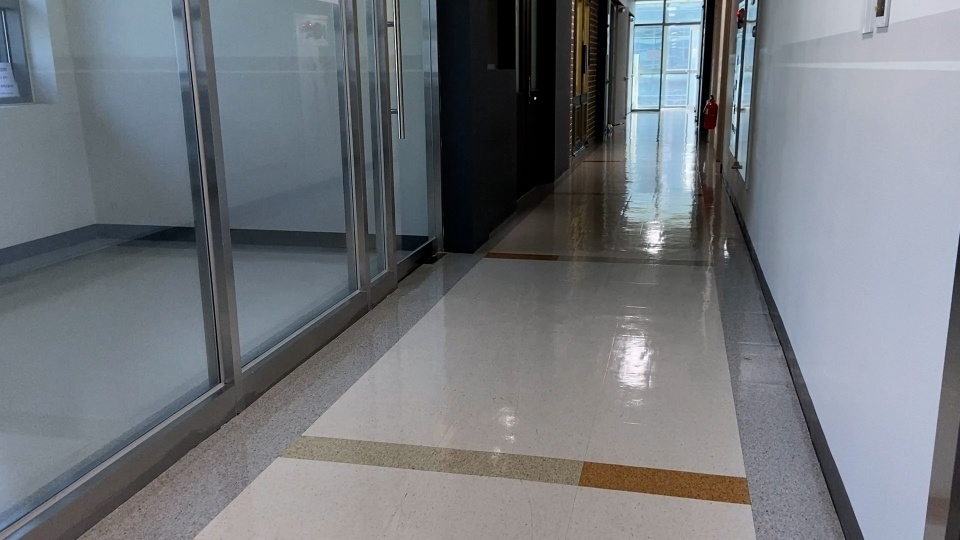}
    &
    \includegraphics[
      width=\colw,
      height=0.285\textheight,
      keepaspectratio
    ]{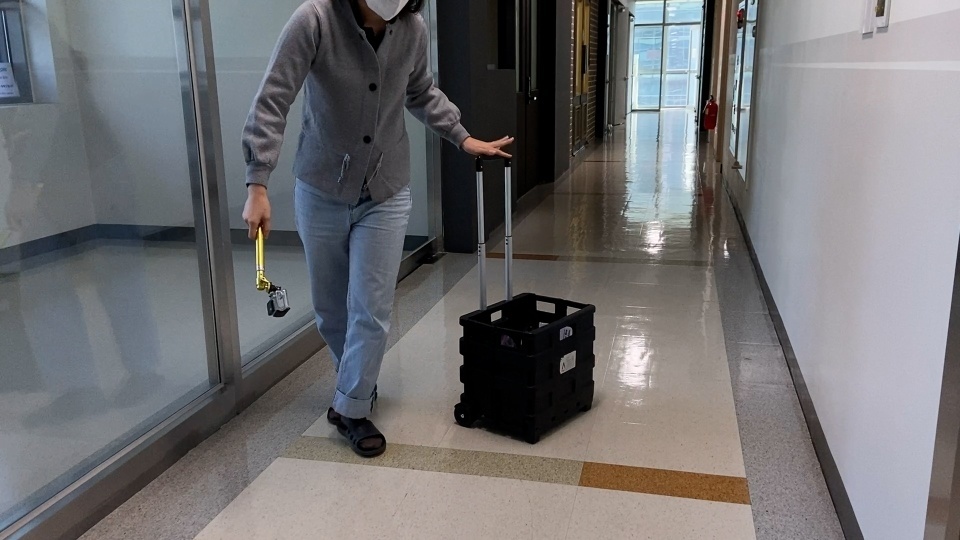}
    \\[4pt]

    \multicolumn{2}{@{}p{\fullw}@{}}{%
      \centering
      \colorbox{black!5}{%
        \parbox[t]{\dimexpr\linewidth-2\fboxsep\relax}{%
          \centering\scriptsize\sffamily
          \textbf{Instruction:}
          Add the woman, wearing a gray jacket and jeans, who is transporting a black rolling crate, holds a yellow and black object in her hand. The hallway, with its industrial lighting and cream-colored walls, stretches out behind her.
        }%
      }%
    }%
  \end{tabular}

  \caption{Example of a triplet obtained from the RORD dataset.}
  \label{fig:rord}
\end{figure*}

\begin{figure*}[t]
  \centering

  \setlength{\tabcolsep}{0pt}

  \begin{tabular}{@{}c@{\hspace{6pt}}c@{\hspace{8pt}}c@{\hspace{6pt}}c@{}}
    \includegraphics[width=\dimexpr0.25\textwidth-5pt\relax,height=\imgheight,keepaspectratio]
      {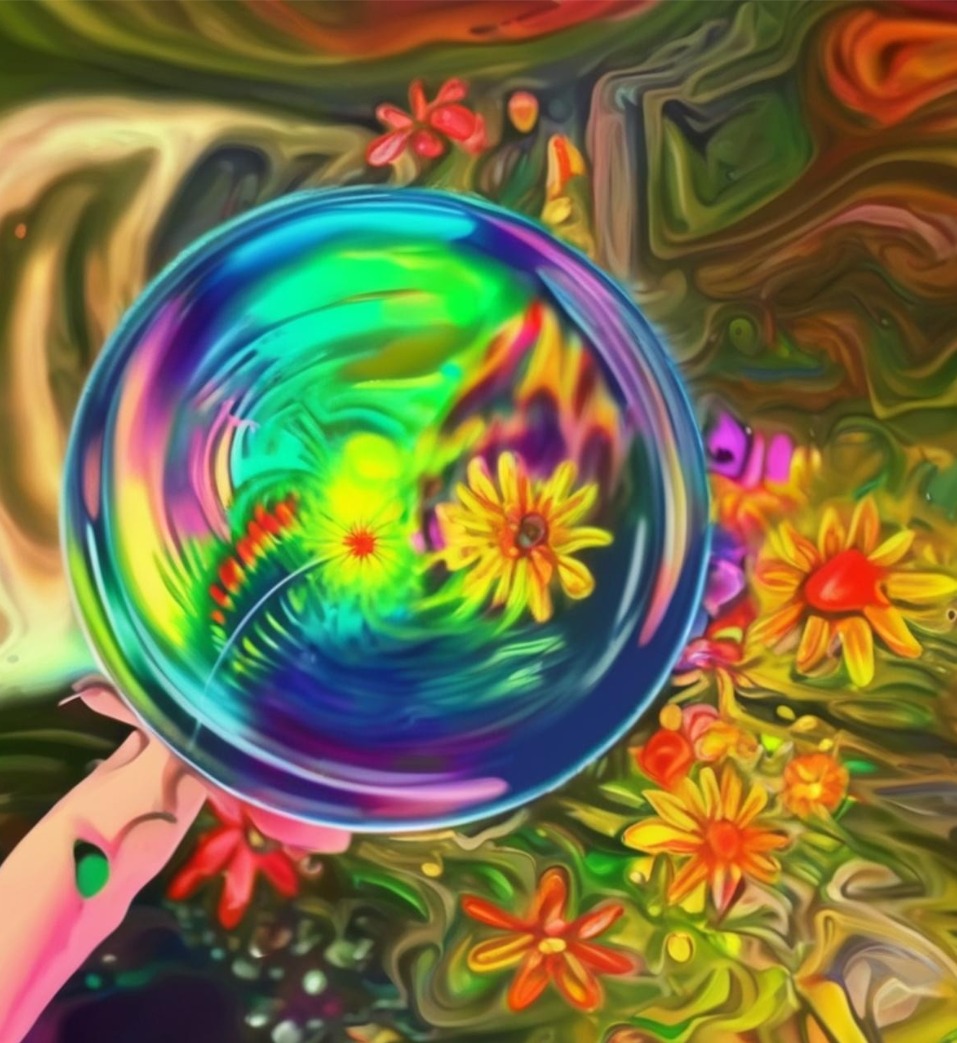} &
    \includegraphics[width=\dimexpr0.25\textwidth-5pt\relax,height=\imgheight,keepaspectratio]
      {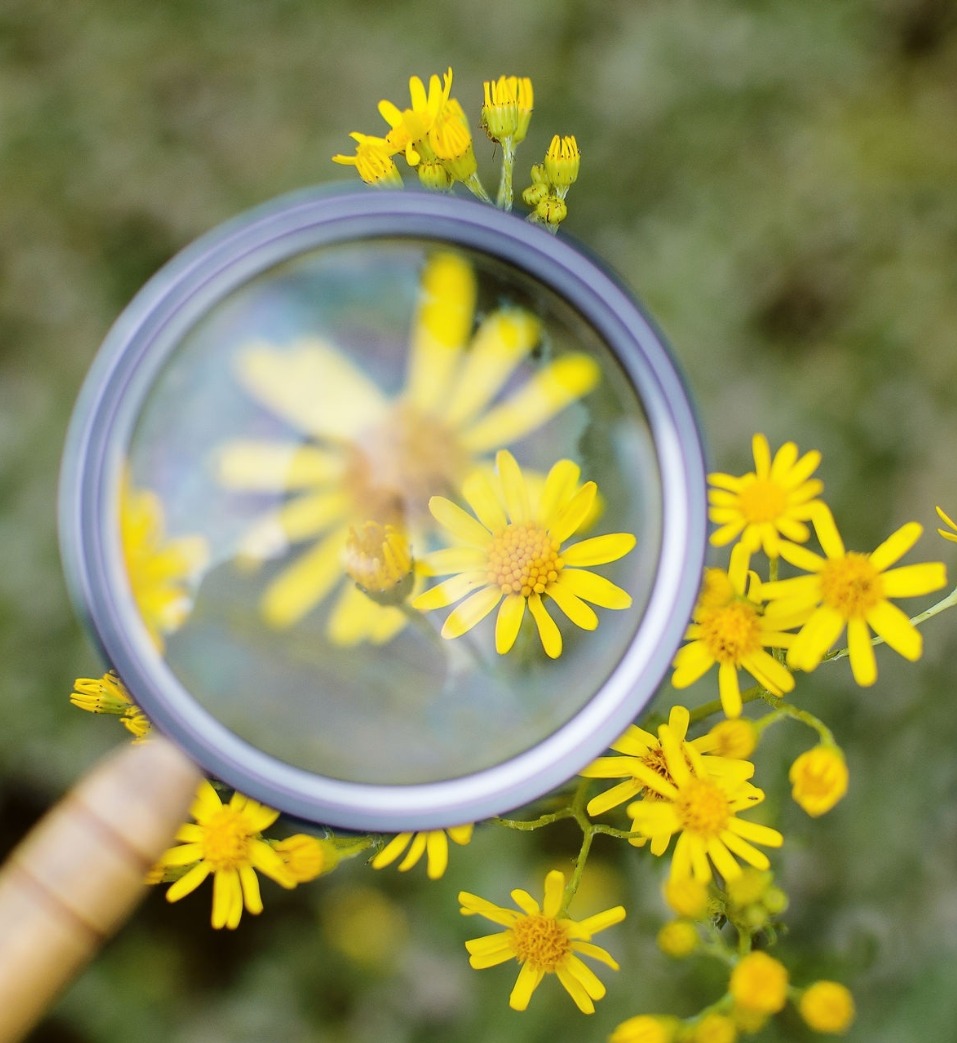} &
    \includegraphics[width=\dimexpr0.25\textwidth-5pt\relax,height=\imgheight,keepaspectratio]
      {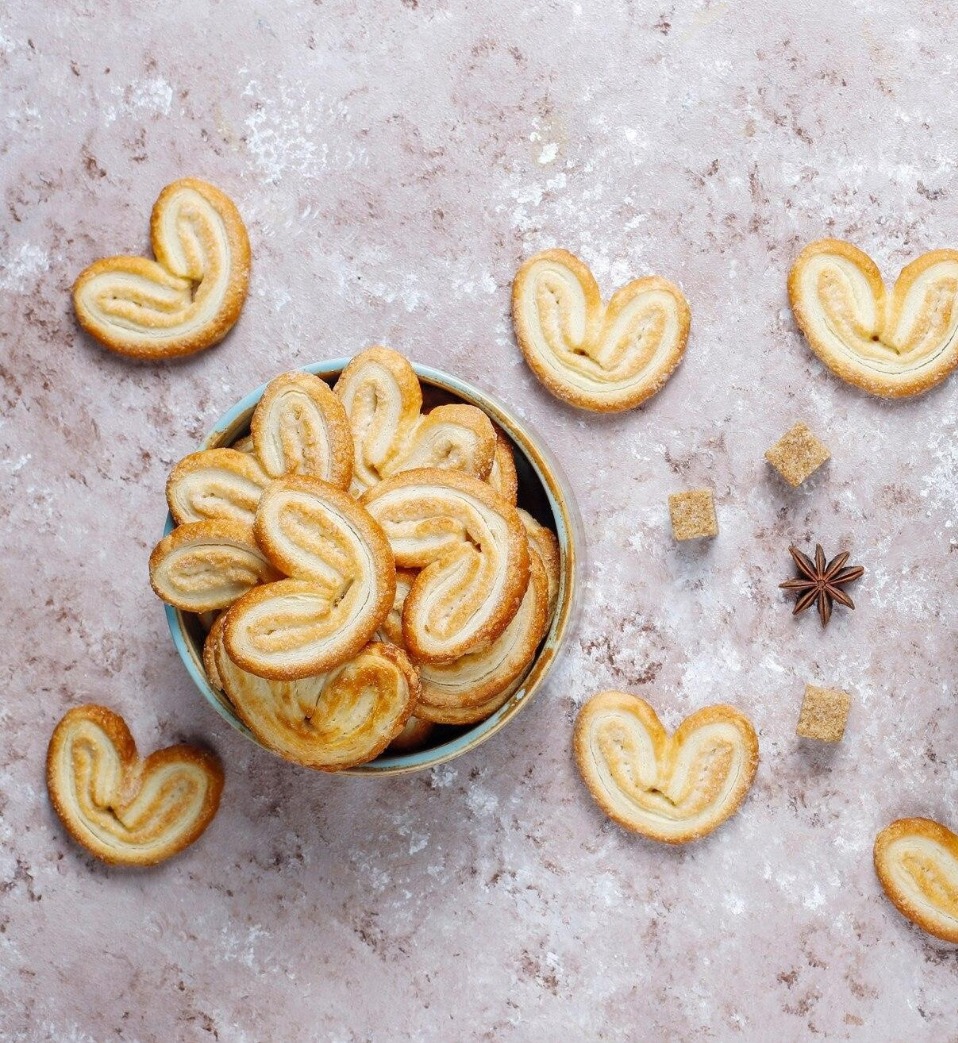} &
    \includegraphics[width=\dimexpr0.25\textwidth-5pt\relax,height=\imgheight,keepaspectratio]
      {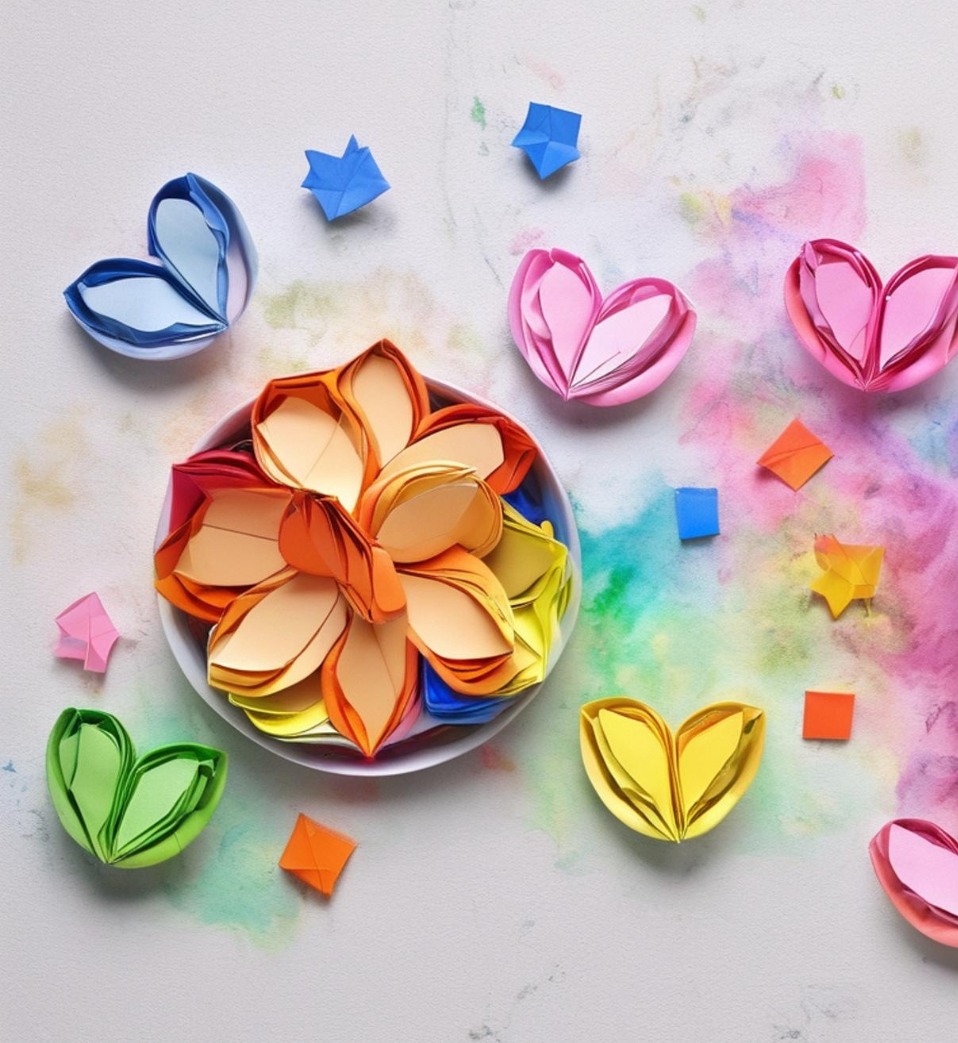} \\
    \noalign{\vspace{2pt}}
    \noalign{\vspace{1pt}}
    \multicolumn{2}{@{}p{\dimexpr0.5\textwidth-4pt\relax}@{}}{%
      \centering
      \colorbox{black!5}{%
        \parbox{\dimexpr0.5\textwidth-4pt-2\fboxsep\relax}{%
          \centering\scriptsize\sffamily\textbf{Instruction:} Transform  style to real%
        }%
      }%
    } &
    \multicolumn{2}{@{}p{\dimexpr0.5\textwidth-4pt\relax}@{}}{%
      \centering
      \colorbox{black!5}{%
        \parbox{\dimexpr0.5\textwidth-4pt-2\fboxsep\relax}{%
          \centering\scriptsize\sffamily\textbf{Instruction:} Stylize the image in a colorful origami style%
        }%
      }%
    } \\
  \end{tabular}

  \vspace{10pt}

  \setlength{\tabcolsep}{0pt}
  \renewcommand{\arraystretch}{1}
  \begin{tabular}{@{}c@{\hspace{8pt}}c@{}}
    \includegraphics[width=\dimexpr0.5\textwidth-4pt\relax]
      {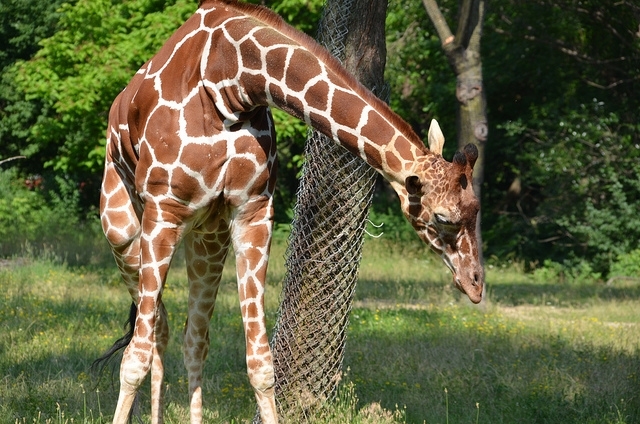} &
    \includegraphics[width=\dimexpr0.5\textwidth-4pt\relax]
      {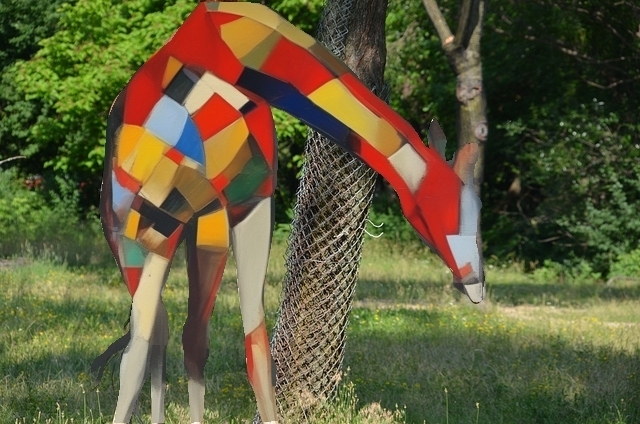} \\
    [4pt]
    \multicolumn{2}{@{}p{\textwidth}@{}}{%
      \centering
      \colorbox{black!5}{%
        \parbox[t]{\dimexpr\textwidth-2\fboxsep\relax}{%
          \centering\scriptsize\sffamily\textbf{Instruction:}
          Draw a giraffe in a cubist style%
        }%
      }%
    }%
  \end{tabular}

  \caption{Examples from the LVIS stylization dataset.}
  \label{fig:stylization_examples}
\end{figure*}

\subsection{Supervised Fine-Tuning datasets}
\label{sec:datasets:sft}
In this section, several of the most novel and practically important approaches used to mine triplets for the SFT stage are described.

\begin{table}[t]
\centering
\caption{Principal triplet sources after filtering. LVIS, HaGRID, and EasyPortrait contribute to both stages; only a subset of their samples is used during pretraining.}
\renewcommand{\arraystretch}{1.05}
\begin{tabular}{@{}p{0.72\linewidth} r@{}}
\hline
\multicolumn{2}{@{}l@{}}{\textbf{Pretraining}} \\
\hline
UltraEdit Remake & \num{6420724} \\
Aurora & \num{160373} \\
LVIS & \num{1000000} \\
HaGRID & \num{107619} \\
EasyPortrait & \num{40000} \\
\hline
Total & $\approx \num{7728776}$ \\
\hline
\multicolumn{2}{@{}l@{}}{\textbf{Supervised fine-tuning (SFT)}} \\
\hline
Autonomous self-mining pipelines & \num{2913829} \\
LVIS & \num{1000000} \\
NHR-Edit & \num{720088} \\
Stylization & \num{726560} \\
Concept Sliders & \num{195525} \\
SEEDPS (parts 2 and 3) & \num{189572} \\
Automated inpainting & \num{177739} \\
HaGRID & \num{107619} \\
EasyPortrait & \num{40000} \\
GIER & \num{5462} \\
Low-level processing dataset & \num{3597} \\
Real tripod photos & \num{4139} \\
Other sources (manual in-house retouching, manual inpainting, 3D renders, and smaller curated collections) & $\approx \num{800000}$ \\
\hline
Total & $\approx \num{6800000}$ \\
\hline
\textbf{GAR based Dataset} &  \\
Total & \num{176532} \\
\hline
\end{tabular}
\label{table:datasets}
\end{table}

\subsubsection{Real Tripod Photos}
A substantial limitation of most automated mining methods is that either the input or the output image contains synthetic artifacts that can bias training. Because the target is pixel-accurate editing and physical plausibility (e.g., shadows, reflections, transparent materials), real triplets with strict camera immobility were additionally collected.

Prior work (e.g., ObjectDrop~\cite{objectdrop} and OmniPaint~\cite{omnipaint}) suggests that even a few thousand high-quality real pairs can substantially improve the modeling of object-induced effects such as shadows and reflections. Motivated by this, a crowdsourcing platform was used with a task that required capturing a ``before'' and ``after'' photo under a strict no-shift protocol (tripod or equivalent locking method). Detailed user instructions described what can and cannot be photographed. In total, \num{4139} triplets (including augmentations) were collected. See Example XXX.

\subsubsection{Real Triplets from Videos}
To obtain more triplets of comparable physical realism, the existing RORD dataset~\cite{RORD} was leveraged, which consists of frames extracted from videos recorded with a static camera. Only indoor scenes were used, since outdoor videos often contain small background changes (e.g., pedestrians, cars, moving leaves, animated billboards, or traffic lights) that violate the no-shift requirement.

Because the videos contain many near-duplicate frames, only about 10\% were retained. Two selection strategies were evaluated: sampling I-frames from MP4 files, and selecting diverse frames using MobileNetV3 embeddings~\cite{mobilenetv3} via~\cite{imagededup}; the embedding-based approach performed better. For person addition, it was required that the target image included at least the upper body and a fully visible head (not partial body parts). Samples were filtered with Qwen2-VL~\cite{qwen2_vl} to enforce this constraint.

Multiple blur measures (Variance of Laplacian, FFT-based metrics, and the Tenengrad measure) were evaluated, and the Blur Effect metric~\cite{blur_effect} (implemented in~\cite{scikitimage}) worked best. Finally, Qwen2-VL~\cite{qwen2_vl} was used to generate editing instructions. See an example of a dataset triplet in Figure~\ref{fig:rord}.

\subsubsection{Virtual Try-On}
The VITON-HD dataset~\cite{vitonhd} was processed with OOT-Diffusion~\cite{ootdiffusion} to obtain paired examples for garment changes. To minimize artifacts, only images where the person’s hands and hair do not overlap the clothing were kept.
To make the resulting triplets more realistic and diverse, a set of background images was collected and the person was composited onto them at several positions and scales. Person mattes were extracted with StyleMatte~\cite{stylematte}, then both the original VITON images and the OOT-Diffusion outputs were composited onto the same backgrounds. After compositing, the images were harmonized with DucoNet~\cite{duconet} to make the lighting more consistent.
To generate instructions, the target garment image (without the person) from VITON-HD was first captioned using LLaMA-3.2-Vision-Instruct-11B~\cite{llama32}, and then rewritten into editing instructions of varying lengths using LLaMA-3.1-8B-Instruct~\cite{llama31}.
Despite these steps, some artifacts remained, e.g., mismatched skin tone, missing or distorted tattoos, neck and jewelry inconsistencies, sleeve artifacts, and matting issues such as white halos around the subject or coarse masks, so a final filtering stage with an assessor was applied.

\subsubsection{Stylization}
The stylization dataset was composed of 2 parts:

Object-level stylization. The LVIS dataset~\cite{lvis} and its instance segmentation annotations were used to stylize only selected objects in an image. To our knowledge, there is no existing dataset for this setting, although the task is challenging and highly relevant for real-world applications. The entire image was first stylized using Stable Diffusion XL~\cite{sdxl} with a Depth ControlNet~\cite{controlnet}, and then the original image was composited with the stylized object region using the LVIS mask.

Full-image stylization. Images from LAION~\cite{schuhmann2022laionb} were stylized using SDXL with a Depth ControlNet, and images from Open Images Dataset v7~\cite{OpenImagesv7} were stylized using Qwen-Image~\cite{qwen_image}. To enable an additional capability, these triplets were inverted to obtain the task ``change any style to realistic.''

Overall, the dataset covered more than 500 styles and contained \num{363280} stylized triplets, along with the same number of inverted triplets. See Figure~\ref{fig:stylization_examples} for the examples.

\subsubsection{Visual Concept Sliders}~\cite{concept_sliders} provide fine-grained attribute control in diffusion models with LoRA adaptors~\cite{lora}. Using this approach, base images were generated with SDXL~\cite{sdxl} and paired edits were produced that modify a single attribute, with controllable intensity and direction when supported by the slider.

To reduce ambiguous cases, prompts were crafted that discourage the attribute from being already shifted (e.g., ``A man of medium build is standing in the center of the square...'' for a muscularity slider), then controlled variations were generated.

Using this approach, the following slider categories were mined: \\
\textbf{Surprised}: controls the degree of surprise. \\
\textbf{Age}: increases apparent age. Only the positive (aging) direction was used. \\
\textbf{Chubby}: controls perceived chubbiness. \\
\textbf{Muscular}: controls muscularity. It was applied to both people and generated animals, where it worked unexpectedly well. \\
\textbf{Tropical}: controls perceived ``tropicalness'' of a scene. This slider performed poorly on average, so prompts were restricted to scenes where the concept is visually supported (e.g., forested environments). \\

To enrich and standardize instructions, MiVOLOv2~\cite{mivolo,beyond_specialization} was used for age and gender estimation. This enabled instructions such as ``Create an image of this \{gender\} at \{age\} years'' with explicit age values. All images were required to contain exactly one person, enforced using a detector model.

Because slider-based edits can unintentionally alter non-target attributes, additional constraints were applied using age and gender estimation. Gender preservation was enforced for all sliders, and for non-age sliders the age change was limited to at most 3 years. Fixed-seed generations were also used to create additional transitions between sliders, e.g., an original image, a ``surprised'' variant, and a ``smiling'' variant from the same seed can yield an instruction like ``Make surprised \{gender\} smile a little''.

See Figure~\ref{fig:sliders_example}. Using this method, \num{195525} triplets were mined.

\subsubsection{Autonomous triplet-mining pipelines}
Multiple configurations of the self-mining pipeline from~\cite{nhr} described in Section~\ref{section:datasets:pretrain:self_data} were used. While configurations differed in the generator stack and filtering stages, they shared the same high-level structure: over-generate candidate edits and retain only those that pass automated validation.

For SFT, diverse input sources were used including Open Images Dataset v7~\cite{OpenImagesv7}, multiple open-source collections of real photos, and images scraped from a range of internet domains, with an emphasis on realistic user-like photography.

As generative models, Qwen-Image~\cite{qwen_image} and proprietary models were used. Using this method, \num{2913829} triplets were mined, including additional filtering described in Section~\ref{section:datasets:filtering} and the same augmentation techniques.

\subsubsection{Automated Inpaint}
Inpainting triplets were generated using inpainting-capable diffusion models with ControlNet conditioning~\cite{controlnet}. Combined with LVIS and Alimama datasets that include segmentation annotations, this yielded \num{177739} triplets. See Figure~\ref{fig:lvis_inpaint} for the example.

\subsubsection{Perception and Recognition Datasets}
For pretraining, and a smaller portion mixed into SFT, several computer vision and perception datasets were incorporated to strengthen base visual understanding, with an emphasis on human body and face anatomy, as well as object localization.

\textbf{HaGRID}~\cite{HAGRID} was used to construct instruction-based triplets by inpainting gestures within annotated bounding boxes and generating prompts such as ``add gesture X''. Using this procedure, \num{107619} triplets were mined.

For facial anatomy, \textbf{EasyPortrait}~\cite{EasyPortrait} was used: selected face parts were masked and then inpainted, yielding \num{40000} triplets.

Finally, \textbf{LVIS} was used to generate segmentation-centric triplets. One or more objects were sampled from LVIS annotations and instructions were produced that require localizing and segmenting these objects. Background-removal triplets were also created where the model is instructed to remove the background and all objects except one (or a small set) of selected instances, resulting in \num{1000000} triplets. See Figure~\ref{fig:lvis_back_example} for an example.

\subsubsection{Open Source Datasets}
For pretraining, the Aurora~\cite{aurora} dataset (\num{160373} triplets) was also used.

For SFT, the following open-source datasets were used (filtered or augmented depending on quality and whether multi-turn edits were available): SEEDPS~\cite{ge2024seed} (parts 2 and 3) (\num{189572} triplets), GIER~\cite{GIER} (\num{5462} triplets), low-level-processing dataset~\cite{LLPD} (\num{3597} triplets), and NHR-Edit~\cite{nhr} (\num{720088} triplets).

\subsection{Generation Augmented Retrieval based Dataset}
\label{sec:dataset_log_grounded}

\paragraph{Data format and motivation}
There are several common ways to build datasets for image editing:

\begin{itemize}
  \item \textbf{Fully synthetic (automatic).}
  Captions and edit instructions are generated by an LLM; the input image is synthesized with a text-to-image model; the edited image is produced by either a specialized module (e.g., inpainting, sliders) or a general-purpose image editor.
  This approach scales well, but is highly prone to domain shift.

  \item \textbf{Semi-synthetic (automatic).}
  The same pipeline, but the input image is a real photograph rather than a generated one.

  \item \textbf{Semi-synthetic (manual).}
  A human writes the instruction and a professional artist performs the edit.
  This typically yields much higher quality, but is expensive and hard to scale.

  \item \textbf{Fully real (manual).}
  Both images are real photos captured under controlled conditions (e.g., using a tripod/locked camera), which best preserves pixel-level alignment and faithfully captures lighting, shadows, reflections, and transparency.
  However, it is difficult to scale and limited to a constrained set of edits (e.g., it cannot cover stylization or adding fantastical objects).

  \item \textbf{Real images (automatic).}
  Triplets mined from videos.
  This can scale well (especially if instructions are generated), but extracting consistent pairs is challenging when the camera or the scene is dynamic.
\end{itemize}

Across these settings, instructions are often treated as ``real'' if they are written by a person. In practice, however, asking annotators to invent edit instructions for dataset creation does not match how image editing models are used. If the text distribution is expected to reflect real-world behavior, it requires genuine user edit queries.

One option is to use Photoshop request datasets, but these are typically small. Another is to reuse prompts from model-comparison platforms (e.g., diffusion ``arenas''), similar to what is done for text-to-image in the Open Image Preferences dataset. However, this source is still biased: the prompts are written for anonymous model ranking rather than natural user intent. Even large-scale prompt collections like DiffusionDB (from Discord) tend to over-represent experienced prompt writers and include keyword-heavy phrasing that is unnatural as everyday language

Therefore, we collected real-world requests from all available open and internal sources, cleaned this corpus to remove noise and non-edit intents, removed duplicates, and made all other necessary preparations.

\paragraph{Image Data Sources}
Open Images V7~\cite{kuznetsova2020open} was used as the source image dataset and $\sim$200k samples were downloaded at 2K resolution, which served as anchor images for the editing queries. To model natural instruction language, a set of in-the-wild editing instructions was used.

\paragraph{Discovering an edit taxonomy}
To identify the most common user intents, the collected corpus was clustered. First, each instruction was embedded using the FRIDA embedder~\cite{snegirev2025russian}. Next, clustering yielded 50 large clusters that correspond to stable semantic groups of requests. Finally, Qwen3-VL-32B~\cite{qwen3vl2025} was used to interpret each cluster by generating a human-readable category name and cluster description. The category list was further expanded with a small set of heuristic additions to improve coverage for rare but important edit modes.

The result of this stage is a practical taxonomy of edits that corresponds to the actual distribution of user requests and is suitable as a "basis" for subsequent instruction generation.

\paragraph{Image-Conditioned Instruction Generation}
Using the discovered semantic clusters, an initial set of image-conditioned edit instructions was synthesized for downloaded samples from the Open Images dataset. For each image, 8 instructions were generated that span different categories in the taxonomy, using Qwen3-VL for generation. This stage yielded instructions that were semantically valid and diverse across edit types, but the phrasing could still be noticeably synthetic and might require further grounding to better match real user language.

\paragraph{Retrieval-based grounding to real user phrasing}
To replace synthetic wording with natural user language, a retrieval pipeline was built over FRIDA embeddings in paraphrase mode.

All unique instructions were indexed in a Qdrant~\cite{qdrant} vector database. For each artificial instruction $Q_\text{art}$, its embedding was computed and the top-K nearest instructions were retrieved (with $K{=}20$).
One user intent $Q_\text{user}$ was then selected via stochastic sampling from the top-K candidates, converting similarities into a probability distribution using a softmax:
\begin{equation}
    p_i=\frac{\exp(s_i)}{\sum_{j=1}^{K}\exp(s_j)},
\end{equation}
where $s_i$ is the cosine similarity score between $Q_\text{art}$ and the $i$-th $Q_\text{user}$. This choice (instead of deterministic top-1) increases lexical diversity and reduces overuse of the same "ideal" formulations.  

\paragraph{Mitigating bias toward popular prompts}
Pairs with insufficient semantic similarity were filtered out using a threshold criterion, limiting the risk of semantic drift when replacing text.  

A nearly inevitable effect of retrieval matching is concentration on a small set of well-phrased instructions that match many artificial instructions. To preserve diversity, a frequency cap was introduced: the same instruction may appear in the final dataset at most 3 times.

\paragraph{Validating instruction applicability to the image}
Even with strong semantic similarity to the synthetic intent, mismatches with the image can occur (e.g., the instruction refers to an object absent from the scene). Therefore, a VLM-based validation stage was added: Gemini 3 Flash checked whether the instruction is applicable to the image $x$. If the instruction was not applicable, the model attempted to apply a \emph{minimal} text edit (preserving the original style) to make the instruction executable for the given image; if minimal correction was not possible, the pair was discarded. This step acts as an “instruction $\leftrightarrow$ image consistency” filter and a gentle correction mechanism for borderline cases.

\paragraph{Generating target images}
After filtering and deduplication, we obtained $\sim$10k images, each associated with 4 to 8 valid in-the-wild instructions. To produce target images $y$, pairs $(x,\;t)$ were sent to high-capacity proprietary image editing models, and the edited results were collected to form final triplets $(x,\;t,\;y)$. We distributed the workload across several models to balance quality and diversity. The editing results were filtered using an in-house Qwen2.5-VL assessor (a detailed model description is provided in Sec.~\ref{sec:assessor}).

\begin{figure}[t]
    \centering
    \includegraphics[width=0.95\columnwidth]{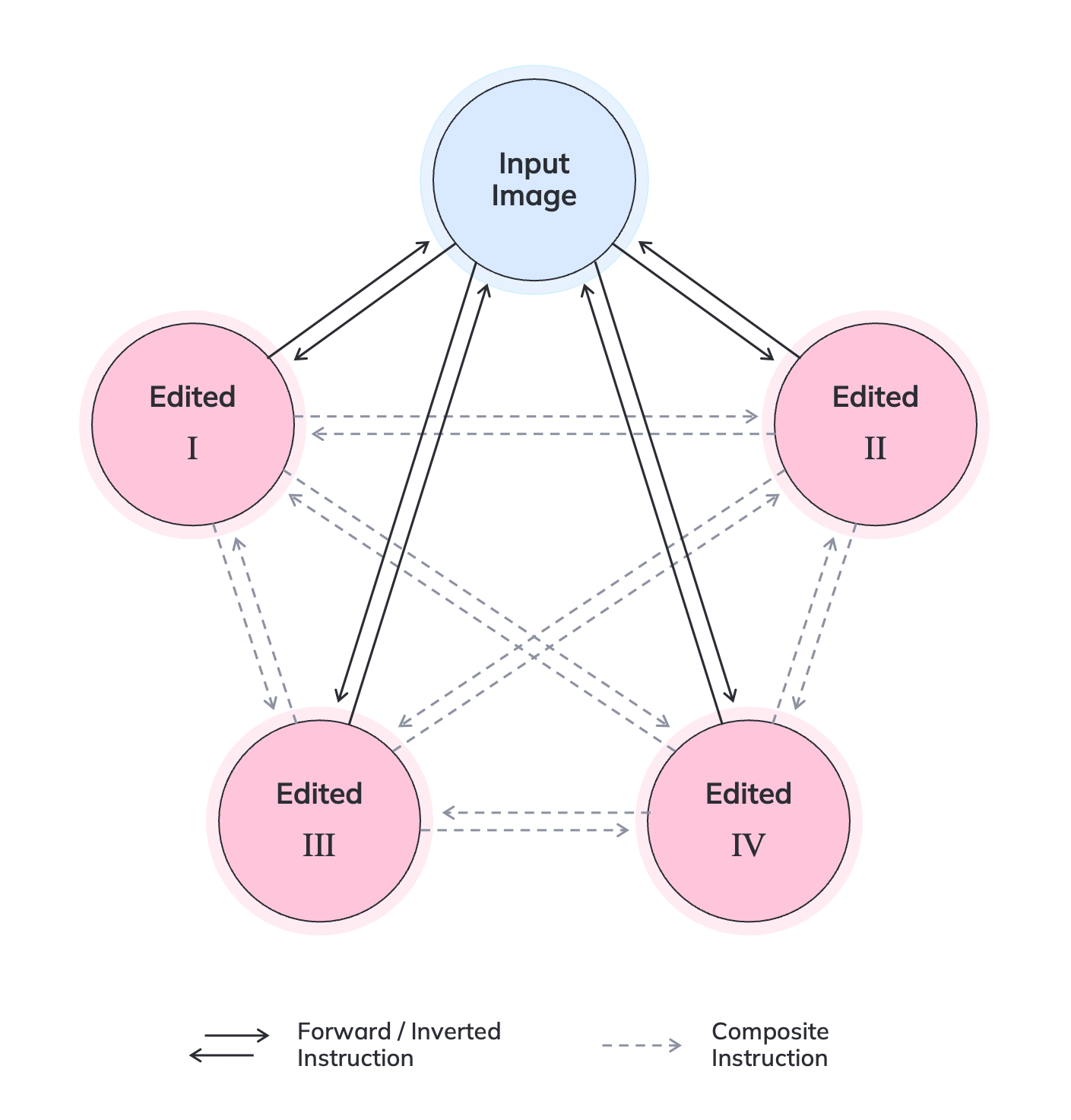}
    \caption{Composite mining process.}
    \label{fig:composite}
\end{figure}

\paragraph{Inverted and composite instructions.}
After generating the target images $y$, the number of training triplets was further increased by reusing already generated edits for the same source image. A visual illustration is provided in Figure~\ref{fig:composite}.

Assume that for an input image $x$ we obtain $N$ edited variants $\{y_i\}_{i=1}^{N}$ with corresponding instructions $\{t_i\}_{i=1}^{N}$, where each mapping $(x,t_i)\mapsto y_i$ forms a base triplet $(x,t_i,y_i)$.

\emph{Instruction inversion.}
For each edited image $y_i$, the reverse editing task is constructed: recover the original image $x$ from $y_i$.
This corresponds to building an inverse instruction $t_i^{-1}$ describing the transformation
\[
(y_i,\;t_i^{-1})\mapsto x,
\]
which yields additional triplets of the form $(y_i,\;t_i^{-1},\;x)$ and thus makes the dataset bidirectional with respect to the source scene.

\emph{Composite transitions between two edits.}
In addition, using the shared “anchor” $x$, transitions between pairs of edited images $(y_i, y_j)$ for $i\neq j$ are constructed.
Intuitively, to move from $y_i$ to $y_j$, one needs (i) to undo the edit that produced $y_i$ (i.e., apply $t_i^{-1}$), and then (ii) apply the edit $t_j$.
A composite instruction $t_{i\rightarrow j}$ was formed that is semantically equivalent to the sequence $(t_i^{-1}\ \text{then}\ t_j)$ and corresponds to the transformation
\[
(y_i,\;t_{i\rightarrow j})\mapsto y_j.
\]
Therefore, for a fixed source image $x$ and a set of $N$ edits, the number of possible directed transitions between edited variants is $N(N-1)$.

Resulting dataset integrated (i) wide coverage of scenes and object categories from Open Images V7, (ii) user-like instruction phrasing obtained by grounding synthetic intents in large-scale in-the-wild queries, (iii) explicit instruction-image consistency enforcement via a VLM-based applicability filter, and (iv) a standardized pipeline for generating target edits.

The final dataset comprised \num{176532} triplets.

\subsection{Issues and Filtering}
\label{section:datasets:filtering}
A task-tuned Gemini validator~\cite{gemini} from~\cite{nhr} was initially employed to clean all SFT datasets, covering forward, backward, and bootstrapped operations. A filtering threshold of $3.5$ was applied, resulting in the removal of approximately $15\%$ of the data. Visual inspection of the retained samples confirmed that this metric effectively preserved high-quality instruction alignment.

These issues were addressed directly. The diffusion-based editing models occasionally produced high-frequency artifacts resembling checkerboard patterns or JPEG compression noise at the borders of outpainting regions and in random parts of the image, particularly on human faces and uniform regions like the sky. Empirical analysis revealed a strong correlation between these visual artifacts and spatial shifts, most notably the repositioning of human faces from their positions in the input images.

\begin{figure}[t]
    \centering
    \includegraphics[width=0.95\columnwidth]{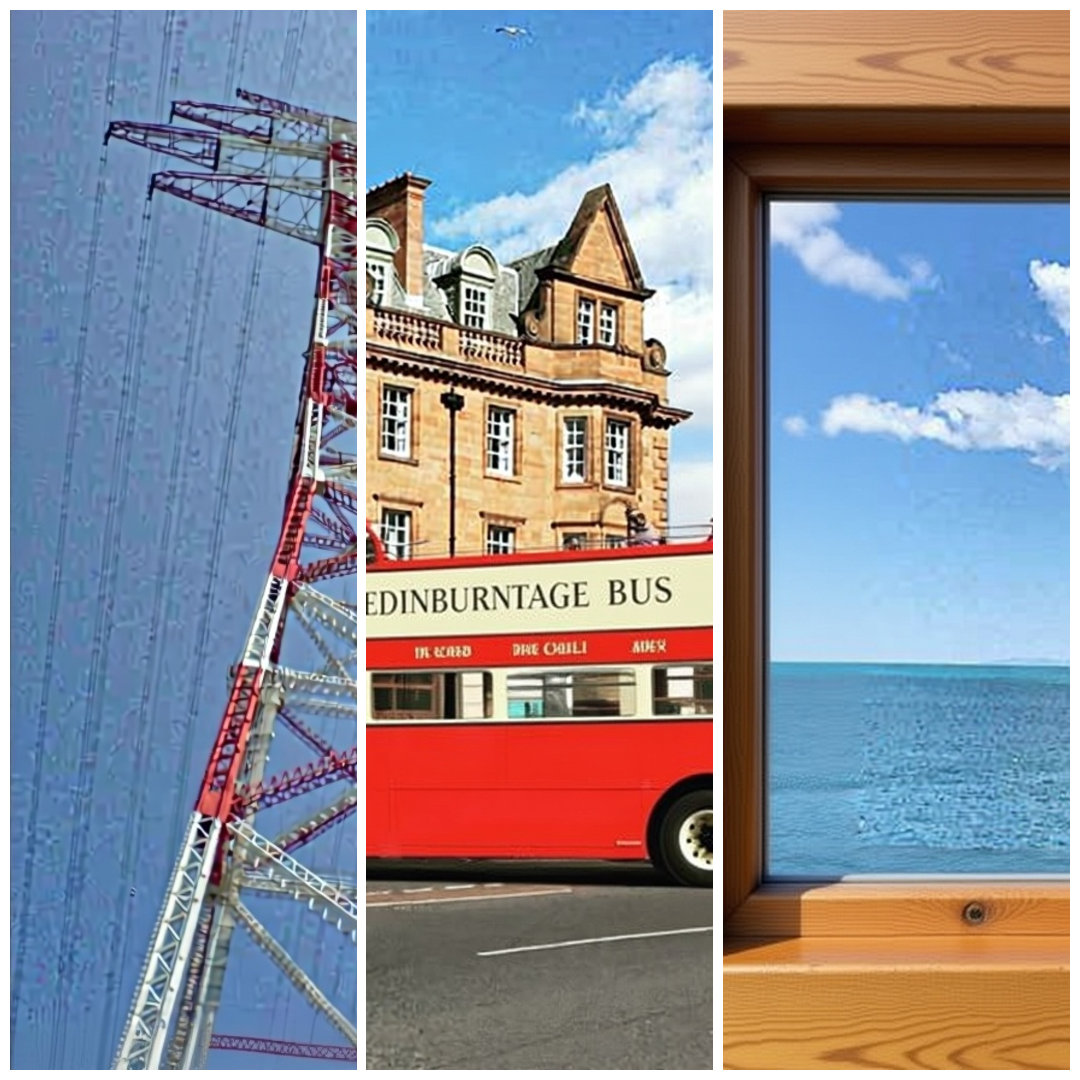}
    \caption{Cases of artifacts on edited images.}
    \label{fig:spoiled_images}
\end{figure}

Since standard detection algorithms proved ineffective, a geometric filtering heuristic based on facial alignment was implemented. For each input-output pair, faces were detected and the Intersection over Union (IoU) of the largest detected face was calculated. A strict spatial constraint was enforced, discarding any training pairs where the face IoU fell below a threshold of $0.9$. While this aggressive filtering resulted in the removal of approximately $35\%$ of the data, it proved essential for eliminating the visual artifacts and preventing the model from memorizing these degradation patterns.

Additionally, both generated and real data (predominantly generated samples) were found to frequently suffer from minor global geometric inconsistencies, such as small shifts, unintended crops, or stretching. To mitigate this, the homography between the input and output images was calculated to align the pairs precisely. This correction ensured spatial consistency, allowing the model to focus on the editing task rather than compensating for trivial misalignment.

\subsection{Synthetic Augmentation Pipeline}
To robustly adapt the model to varied user inputs, a \textbf{Just-in-Time (JIT)} synthetic augmentation strategy was employed. Instead of generating static files, images from the prepared dataset were dynamically transformed during training to create new triplets on the fly. This effectively multiplied the dataset size and enforced consistency across diverse editing scenarios.

\paragraph{Bidirectional Photometric and Restoration Operations.}
Reversible transformations were grouped into pairs to facilitate bidirectional learning. The model was trained to both apply and reverse effects for \textit{blur/deblur}, \textit{noise/denoise}, \textit{sepia/desepia}, and \textit{grayscale/colorization}. Crucially, for the \textit{colorization} task, the source image was not simply desaturated. Instead, an upgraded grayscale synthesis pipeline was employed that simulated analog film characteristics through randomized channel mixing, sigmoid contrast adjustments, and realistic grain injection. Additionally, scalar adjustments for brightness, contrast, and saturation were employed in both increasing and decreasing directions to cover a full spectrum of global photometric changes.

\paragraph{Instruction Adherence and Invariance.}
To prevent over-editing and ensure strict adherence to prompts, two specific constraints were introduced:
\begin{itemize}
    \item \textbf{Identity Mapping (``Do Not Change''):} Triplets where the source and target images are identical were generated. Paired with passive instructions (e.g., ``do nothing''), this taught the model to preserve image fidelity when no edits are requested.
    \item \textbf{Mirror Augmentation:} Horizontal flipping was selectively applied to inputs to increase visual diversity. Crucially, this was conditional: mirroring was disabled for prompts containing directional terms (e.g., ``left'', ``text'') to ensure the model correctly grounds spatial instructions while becoming invariant to global orientation elsewhere.
\end{itemize}

\paragraph{Structural and Typographic Editing.}
Complex structural changes were simulated by overlaying geometric primitives (synthetic inpainting) or rendering variable text. These were paired with precise instructions to ``fill'' areas or modify specific words, training fine-grained spatial control.

\paragraph{Real-world Quality Adaptation.}
To bridge the gap between pristine training data and potential low-quality user uploads, synchronized JPEG compression was applied to both source and target images. This accustomed the model to processing inputs with high-frequency loss and compression artifacts without editing degradation.

\subsection{DPO Data Preparation}
\label{section:dpo_dataset}
To effectively align the model with human preferences and ensure robustness across different editing scenarios, a composite preference dataset $\mathcal{D}_{\text{DPO}}$ was constructed. This dataset was derived from three distinct sources, each targeting specific aspects of generation quality:

\begin{enumerate}

    \item \textbf{Self-Generated Preferences (On-Policy).} 
    A dataset was constructed by generating a large corpus of images using the SFT model itself. These generations were subsequently annotated by the in-house assessment model described in Section~\ref{sec:assessor}, which assigned scores for both aesthetic quality and instruction adherence. Based on these scores, pairs $(x_w, x_l)$ were formed.
    \\
    This dataset served as a feedback signal for self-correction. By exposing the model to its own failures ($x_l$) versus its successes ($x_w$), it effectively targeted the suppression of model-specific visual artifacts, hallucinations, and distortions that arose during the SFT phase.

    \item \textbf{Symmetric Preference Optimization.}
    Similar to the InstructEngine framework~\cite{lu2025instructengine}, which employs cross-validation alignment for T2I generation, a symmetric preference optimization strategy was adopted for the image editing task. For each input pair $(x, c_1)$, where $x$ is the source image and $c_1$ is the target editing instruction, multiple negative instructions were synthesized. These instructions aimed to perform the same type of editing operation but differed in fine-grained details.
    
    For example, given an original instruction $c_1 = \text{``make the chair wooden''}$, hard negative instructions such as $c_2 = \text{``make the table wooden''}$ (object substitution) or $c_3 = \text{``make the chair wicker''}$ (material substitution) were generated. Images $(y_1, y_2, y_3)$ corresponding to these prompts were generated using the SFT model and filtered using the assessor model (see Section~\ref{sec:assessor}) to ensure semantic consistency. Preference pairs were then constructed symmetrically:
    \begin{itemize}
        \item For the original instruction $c_1$, its corresponding generation $y_1$ was designated as the \textit{winner} ($x_w$), while generations from alternative prompts (e.g., $y_2, y_3$) served as \textit{losers} ($x_l$).
        \item Reciprocally, for any alternative instruction (e.g., $c_2$), its specific generation $y_2$ became the \textit{winner}, while the generation from the original prompt $y_1$ and other variants (e.g., $y_3$) functioned as \textit{losers}.
    \end{itemize}
    This approach ensured that every generated image served as both a positive and a hard negative example depending on the conditioning instruction. Consequently, this strategy improved the instruction-following capabilities of the trained model by forcing it to distinguish between closely related semantic concepts.

    \item \textbf{Distillation from Strong Teachers.}
    To enhance the aesthetic quality of generated images, high-quality data from advanced proprietary models was leveraged. This subset was constructed using the proprietary generations collected in Section~\ref{sec:dataset_log_grounded}. To form preference pairs, these proprietary samples were augmented with corresponding images generated by the SFT model. These SFT generations were evaluated using the in-house assessor model described in Section~\ref{sec:assessor} to facilitate the construction of training pairs.
    \\
    This composite strategy acted as a direct distillation mechanism. By aligning the model with the superior outputs of more complex editors, it explicitly encouraged it to emulate their high visual appeal and artistic quality.
    
\end{enumerate}

\section{Results}
\label{results}
In this section, we further analyzed topics that were not fully discussed in Section~\ref{section:method} and benchmarked our best-performing configuration against current state-of-the-art methods.

\subsection{Ablation Studies}
\label{sec:ablation}

\paragraph{Reference Image Guidance: sequence-wise vs. channel-wise}
We first investigated two strategies for incorporating the reference image: sequence-wise and channel-wise concatenation. Our experiments showed that sequence-wise concatenation consistently outperformed channel-wise concatenation in all benchmarks, with the largest gains observed in the model’s instruction-following abilities. However, sequence-wise concatenation introduced a clear computational overhead because it increased the token sequence length, thereby slowing down inference. With Sana’s linear-complexity attention, the inference time approximately doubled. For DiT-based models with standard quadratic attention, the slowdown was even more pronounced, scaled superlinearly with the increased number of tokens, and often became the primary bottleneck at high resolutions.

This trade-off led to the following practical observation:
\begin{observation}[6]
    We observed consistent gains with sequence-wise guidance in metrics, but the practical gains were often incremental relative to its latency cost. In many cases, similar outcomes could be achieved with the channel-wise variant by re-sampling a few times. In contrast, channel-wise concatenation substantially reduced generation latency, yielding a clear improvement in user experience. Therefore, channel-wise guidance is used in our final high-throughput configuration.
\end{observation}

\paragraph{Textual Guidance: Meta Tokens and VLM Connector Design}
In this section, we analyzed how textual guidance was formed and injected into the diffusion denoising process, focusing on the connector design that bridged the VLM representation space with the diffusion conditioning space. We compared three guidance paradigms.

\textbf{(i) Native text encoder.} As a baseline, we used the diffusion model’s native text encoder, which conditioned generation only on the text prompt and did not explicitly incorporate the input image. This setup was attractive because it required no additional connector and avoided an extra alignment stage. However, it was fundamentally limited by the absence of vision-language reasoning: the instruction could not be interpreted in the context of the reference image, which often led to ambiguous edits, especially for compositional modifications that depend on understanding the scene.

\textbf{(ii) Query-based expansion.} Following the approach proposed in~\cite{koh2023generating}, the VLM produced a compact set of 8 guidance tokens, which were then expanded by a Q-Former connector. The Q-Former was initialized with a set of learnable queries whose size matched the maximum conditioning sequence length expected by the diffusion backbone, allowing it to map a short VLM output into a full-length diffusion conditioning sequence.

\textbf{(iii) Meta-token generation.} Inspired by~\cite{pan2025metaqueries}, we prompted the VLM with a set of meta-tokens and let it generate the full conditioning sequence required by the diffusion model in a single forward pass. To bridge the representation gap between the VLM output space and the diffusion conditioning space, we evaluated connectors of different types and depths, including (i) a standard Transformer encoder and (ii) an ELLA-based connector.

We first tested whether the meta-queries paradigm with a standard encoder consistently outperformed the Q-Former setup. To ensure a fair comparison, we used the same connector depth (four blocks~\cite{koh2023generating}) in both cases. We also compared these results against a baseline trained with a native text-only encoder, without multimodal support.
This comparison led to the following observation:
\begin{observation}[7]
    The meta-queries configuration drastically improved the model’s instruction-following capabilities compared to the Q-Former and native-encoder baselines.
\end{observation}

Next, we evaluated timestep-aware conditioning with ELLA against a standard encoder-based connector.
For each setup, we examined how connector depth affected performance by sweeping the number of layers from 2 to 8.
The depth sweep led to the following observation:
\begin{observation}[8]
For both connector configurations, a depth of four blocks was optimal. Compared to a standard encoder, the ELLA connector yielded only minor improvements that were not consistent across settings.
\end{observation}

\begin{table*}[t]
\caption{Quantitative comparison on ImgEdit~\cite{ye2025imgedit}. ``Overall'' is calculated by averaging all scores across tasks. VIBE achieves top-tier overall performance and leads several core edit categories.}
\centering
\small
\setlength{\tabcolsep}{3pt}
\renewcommand{\arraystretch}{1.05}
\begin{tabular}{l|ccccccccc|c}
\toprule
\textbf{Model} &
\textbf{Add} &
\textbf{Adjust} &
\textbf{Extract} &
\textbf{Replace} &
\textbf{Remove} &
\textbf{Background} &
\textbf{Style} &
\textbf{Hybrid} &
\textbf{Action} &
\textbf{Overall}$\uparrow$ \\
\midrule
Instruct-Pix2Pix~\cite{brooks2023instructpix2pix} & 2.45 & 1.83 & 1.44 & 2.01 & 1.50 & 1.44 & 3.55 & 1.20 & 1.46 & 1.88 \\
MagicBrush~\cite{zhang2023magicbrush} & 2.84 & 1.58 & 1.51 & 1.97 & 1.58 & 1.75 & 2.38 & 1.62 & 1.22 & 1.90 \\
AnyEdit~\cite{yu2025anyedit} & 3.18 & 2.95 & 1.88 & 2.47 & 2.23 & 2.24 & 2.85 & 1.56 & 2.65 & 2.45 \\
UltraEdit~\cite{zhao2024ultraedit} & 3.44 & 2.81 & 2.13 & 2.96 & 1.45 & 2.83 & 3.76 & 1.91 & 2.98 & 2.70 \\
OmniGen~\cite{xiao2025omnigen} & 3.47 & 3.04 & 1.71 & 2.94 & 2.43 & 3.21 & 4.19 & 2.24 & 3.38 & 2.96 \\
ICEdit~\cite{zhang2025context} & 3.58 & 3.39 & 1.73 & 3.15 & 2.93 & 3.08 & 3.84 & 2.04 & 3.68 & 3.05 \\
Step1X-Edit-v1.1~\cite{liu2025step1xedit} & 3.88 & 3.14 & 1.76 & 3.40 & 2.41 & 3.16 & 4.63 & 2.64 & 2.52 & 3.06 \\
BAGEL~\cite{deng2025emerging} & 3.56 & 3.31 & 1.70 & 3.30 & 2.62 & 3.24 & 4.49 & 2.38 & 4.17 & 3.20 \\
UniWorld-V1~\cite{lin2025uniworld} & 3.82 & 3.64 & 2.27 & 3.47 & 3.24 & 2.99 & 4.21 & 2.96 & 2.74 & 3.26 \\
OmniGen2~\cite{wu2025omnigen2} & 3.57 & 3.06 & 1.77 & 3.74 & 3.20 & 3.57 & 4.81 & 2.52 & 4.68 & 3.44 \\
FLUX.1 Kontext [Dev]~\cite{flux_kontext} & \underline{4.12} & 3.80 & 2.04 & 4.22 & 3.09 & 3.97 & \underline{4.51} & 3.35 & \underline{4.25} & 3.71 \\
Z-Image~\cite{zimage2025} & \textbf{4.40} & \underline{4.14} & \textbf{4.30} & \textbf{4.57} & \underline{4.13} & \underline{4.14} & \textbf{4.85} & \textbf{3.63} & \textbf{4.50} & \textbf{4.30} \\
\specialrule{0.9pt}{0pt}{0pt}
\textbf{VIBE} & 3.89 & \textbf{4.22} & \underline{2.90} & \underline{4.34} & \textbf{4.42} & \textbf{4.22} & 4.40 & \underline{3.52} & 2.75 & \underline{3.85} \\
\bottomrule
\end{tabular}
\label{tab:imgedit_gpt41}
\end{table*}

\begin{table}[t]
\centering
\caption{GEdit-Bench-EN~\cite{liu2025step1xedit} (Full set)$\uparrow$: Semantic Consistency (G\_SC), Perceptual Quality (G\_PQ), and Overall Score (G\_O).}
\label{tab:gedit_en}
\small
\setlength{\tabcolsep}{3.5pt}
\renewcommand{\arraystretch}{1.05}
\begin{tabular}{l|ccc}
\toprule
\textbf{Model} & {G\_SC} & {G\_PQ} & {G\_O} \\
\midrule
AnyEdit~\cite{yu2025anyedit}                          & 3.18 & 5.82 & 3.21 \\
Instruct-Pix2Pix~\cite{brooks2023instructpix2pix}     & 3.58 & 5.49 & 3.68 \\
MagicBrush~\cite{zhang2023magicbrush}                 & 4.68 & 5.66 & 4.52 \\
UniWorld-V1~\cite{lin2025uniworld}                    & 4.93 & 7.43 & 4.85 \\
OmniGen~\cite{xiao2025omnigen}                        & 5.96 & 5.89 & 5.06 \\
FLUX.1 Kontext [Dev]~\cite{flux_kontext}              & 6.52 & 7.38 & 6.00 \\
OmniGen2~\cite{wu2025omnigen2}                        & 7.16 & 6.77 & 6.41 \\
BAGEL~\cite{deng2025emerging}                         & 7.36 & 6.83 & 6.52 \\
Step1X-Edit-v1.1~\cite{liu2025step1xedit}                 & 7.66 & \underline{7.35} & \underline{6.97} \\
Z-Image~\cite{zimage2025} & \textbf{8.11} &  \textbf{7.72} &  \textbf{7.57} \\
\specialrule{0.9pt}{0pt}{0pt}
\textbf{VIBE}                                   & \underline{7.91} & 6.33 & 6.81 \\
\bottomrule
\end{tabular}
\end{table}

\subsection{Benchmarks and Metrics}
\label{bench_and_metrics}

We evaluated the final model on GEdit-Bench~\cite{liu2025step1xedit} and ImgEdit-Bench~\cite{ye2025imgedit}, strictly following the authors' official evaluation protocols. We compare against a broad set of leading instruction-based editing systems that are either open-weight or otherwise publicly accessible for controlled benchmarking, including several substantially larger backbones. For GEdit-Bench, we used the VIEScore setup with GPT-4.1~\cite{gpt4} to report Semantic Consistency (SC, 0--10), Perceptual Quality (PQ, 0--10), and Overall (O). For ImgEdit-Bench, we adopted the original authors' protocol: GPT-4.1 was used to score edited images across several criteria, each rated on a 1--5 scale.

\subsection{Comparison with Existing Methods}
\label{sec:sota}

VIBE achieved an overall score of \textbf{3.85} on the \textbf{ImgEdit} benchmark, ranking second among the compared methods in Table~\ref{tab:imgedit_gpt41}, and delivering a distinctly strong editor profile. In particular, VIBE leads multiple core categories that demand strict preservation of the input image, including \textbf{Adjust} (4.22), \textbf{Remove} (4.42), and \textbf{Background} (4.22). It also ranks among the top performers on \textbf{Replace}, \textbf{Extract}, and \textbf{Hybrid} edits, indicating robust instruction grounding across a broad range of operations, despite using a markedly smaller diffusion backbone than several of the strongest baselines in the comparison. We observe that the most challenging cases for VIBE are highly complex, non-local edits, such as \textbf{Action}, that require substantial geometric and compositional changes (Table~\ref{tab:imgedit_gpt41}), which likely benefit from larger, more complex models.

On \textbf{GEdit-Bench-EN}, VIBE achieved an overall score of \textbf{6.81} (Table~\ref{tab:gedit_en}). Notably, the model received the second-highest score for semantic consistency (\textbf{7.91}), demonstrating reliable instruction-following behavior. Although our perceptual quality score (6.33) trails behind systems optimized specifically for visual fidelity, the data suggests this gap is due to fine details and minor artifacts rather than a failure in semantic alignment. Together, ImgEdit and GEdit suggest that VIBE prioritizes faithful, minimally invasive edits over aggressive scene redrawing.
\section{Conclusions}
\label{section:conclusions}

The presented work shows that high-quality instruction-based image editing can be achieved with a relatively small model, with the right design and training setup. A strong but compact 2B VLM is enough to read complex user requests in the context of the input image and provide stable guidance via learnable meta-tokens and a lightweight connector. This work shows that even a 1.6B diffusion backbone can deliver high-quality edits. With channel-wise reference guidance, the pipeline keeps high throughput, fits into 24 GB of GPU memory, and can generate 2K images in about 4 seconds on an NVIDIA H100 in BF16.

We show that stability and strict source consistency come not only from architecture choices, but also from consistent work with training stages and data. The paper uses a four-stage setup: first align the VLM-to-diffusion interface with a text-to-image objective (freezing the backbones), then do large-scale pretraining and SFT with mixed editing and T2I data as an anchor, train in mixed resolution with diverse aspect ratios, and finally apply Diffusion-DPO to improve both instruction following and visual quality, including symmetric hard negatives and distillation from strong complex editors. Data quality is critical here, and real-world triplets are hard to get. Instead of only imitating user prompts, the work grounds synthetic intents to real user phrasing via retrieval over real-world requests, validates instruction applicability to the image, and scales triplets further with inversion and compositional bootstrapping. 

Ultimately, we show that with clean data and a disciplined training recipe, a practical editing system can match or surpass significantly larger models on core tasks, especially those requiring strict preservation of input content. Remaining challenges are concentrated in complex edits requiring major geometric changes, as well as fine-grained visual artifacts that continue to limit perceptual quality.
\section{Limitations}
\label{section:limitations}

Despite strong benchmark results and overall high quality, the model has limited capacity due to its relatively low complexity. Very complex operations can still fail, and some hard aesthetic requests remain unstable. In practice, several categories of real-world photos are harder than generated images, since the in-the-wild domain is much more diverse. The range of capture conditions, from old mobile cameras to professional DSLR setups, makes the problem extremely challenging even for large proprietary systems.

For the same reason, the pipeline tends to be more robust on generated images from modern generators, where the data distribution is closer to the training data. Despite extensive filtering, the generative signal in the input, output, or instruction can still dominate over the real-photo signal.

The main purpose of this model is research. The pipeline relies on pretrained components (VLM and diffusion), and a substantial part of the training data is generated automatically. As with other generative systems, we do not guarantee correct or safe behavior in all situations, and the model may produce incorrect, misleading, or otherwise undesirable outputs. Users are responsible for appropriate use in their own setting, including any required rights and consent, and for any decisions made based on the outputs. We do not commit to providing support, updates, or fixes.

We did not perform a systematic evaluation of bias or fairness. Since the pipeline relies on pretrained components, auxiliary models, large-scale open data, and automatically generated samples, the system may inherit biases from these sources.

Strict source consistency can also be intrinsically difficult for some edit types. Even significantly larger closed systems can fail in these cases, so the presented compact model may drift as well. Finally, the VLM backbone is kept frozen across the whole pipeline to preserve its original knowledge, so the effect of full end-to-end VLM adaptation on final quality is not studied.

\section{Future works}
\label{section:future_works}

A clear next step is to reduce inference cost by distilling the model for fewer diffusion steps and removing CFG. Quantization is also a practical direction to improve throughput and memory footprint, potentially enabling faster inference on lower-end hardware.

Another important direction is to increase the share of real-world signal in training data, in both triplets and validation, to improve robustness on real photos. Stronger adaptation strategies also remain open, including partial or full VLM finetuning, to study the trade-off between preserving general knowledge and improving editing-specific behaviors.
\FloatBarrier
{
  \small
  \bibliographystyle{ieeenat_fullname}
  \bibliography{main}
}

\end{document}